\documentclass[acmsmall]{acmart}
\usepackage{graphicx}
\usepackage{amsmath}
\usepackage{multirow}
\usepackage{diagbox}
\usepackage{enumitem}
\setlist[itemize]{leftmargin=*}

\AtBeginDocument{%
  }

\setcopyright{acmcopyright}
\copyrightyear{2023}
\acmYear{2023}
\acmDOI{https://doi.org/10.1145/3617829}

\acmJournal{TOIS}
\acmVolume{42}
\acmNumber{2}
\acmArticle{49}
\acmMonth{11}

\begin{document}

\title{Policy-driven Knowledge Selection and Response Generation for Document-grounded Dialogue}

\author{Longxuan Ma}
\email{lxma@ir.hit.edu.cn}
\orcid{0000-0003-1431-8485}
\affiliation{%
  \institution{Research Center for Social Computing and Information Retrieval, Harbin Institute of Technology}
  \streetaddress{92, Xidazhi Road, Nangang Qu}
  \city{Harbin}
  \state{Heilongjiang}
  \country{China}
}

\author{Jiapeng Li}
\email{jpli@ir.hit.edu.cn}
\orcid{}
\affiliation{  
  \institution{Research Center for Social Computing and Information Retrieval, Harbin Institute of Technology}
  \streetaddress{92, Xidazhi Road, Nangang Qu}
  \city{Harbin}
  \state{Heilongjiang}
  \country{China}
}

\author{Mingda Li}
\email{mdli@ir.hit.edu.cn}
\orcid{}
\affiliation{%
  \institution{Research Center for Social Computing and Information Retrieval, Harbin Institute of Technology}
  \streetaddress{92, Xidazhi Road, Nangang Qu}
  \city{Harbin}
  \state{Heilongjiang}
  \country{China}
}

\author{Wei-Nan Zhang}
\authornote{Corresponding author.}
\email{wnzhang@ir.hit.edu.cn}
\orcid{0000-0001-5981-4752}
\affiliation{  
  \institution{Research Center for Social Computing and Information Retrieval, Harbin Institute of Technology}
  \streetaddress{92, Xidazhi Road, Nangang Qu}
  \city{Harbin}
  \state{Heilongjiang}
  \country{China}
}

\author{Ting Liu}
\email{tliu@ir.hit.edu.cn}
\orcid{0009-0007-1999-2600}
\affiliation{%
  \institution{Research Center for Social Computing and Information Retrieval, Harbin Institute of Technology}
  \streetaddress{92, Xidazhi Road, Nangang Qu}
  \city{Harbin}
  \state{Heilongjiang}
  \country{China}
}

\renewcommand{\shortauthors}{Trovato et al.}

\begin{abstract}
Document-grounded dialogue (DGD) uses documents as external knowledge for dialogue generation. Correctly understanding the dialogue context is crucial for selecting knowledge from the document and generating proper responses. In this paper, we propose using a dialogue policy to help the dialogue understanding in DGD. Our dialogue policy consists of two kinds of guiding signals: utterance function and topic transfer intent. The utterance function reflects the purpose and style of an utterance, and the topic transfer intent reflects the topic and content of an utterance. We propose a novel framework exploiting our dialogue policy for two core tasks in DGD, namely knowledge selection (KS) and response generation (RG). The framework consists of two modules: the Policy planner leverages policy-aware dialogue representation to select knowledge and predict the policy of the response; the generator uses policy/knowledge-aware dialogue representation for response generation. Our policy-driven model gets state-of-the-art performance on three public benchmarks and we provide a detailed analysis of the experimental results. Our code/data will be released on GitHub.
\end{abstract}



\received{10 March 2023}
\received[revised]{23 June 2023}
\received[accepted]{19 August 2023}

\maketitle

\section{Introduction}
\label{introduction}
\begin{figure*}[t]
\centering
\includegraphics[width=\linewidth]{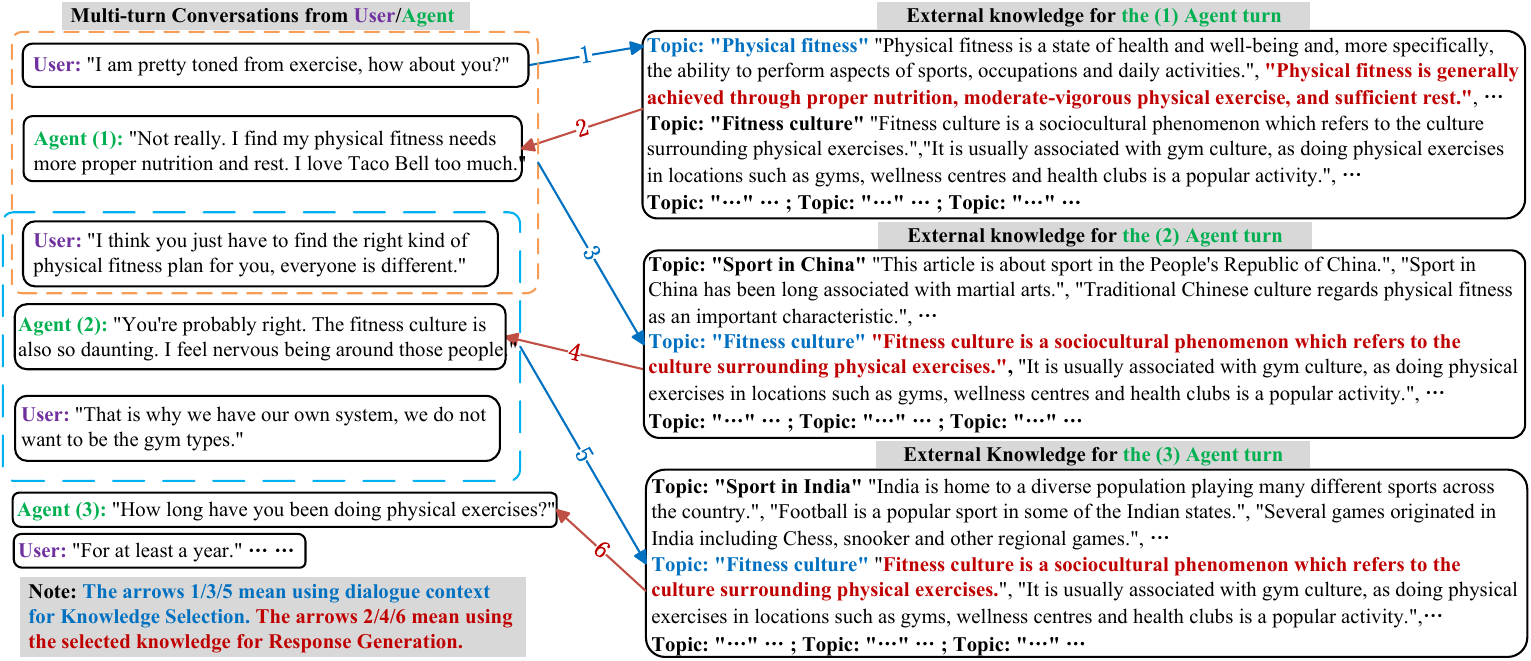}
\caption{A document-grounded dialogue example in the Wizard-of-Wikipedia dataset. For each agent turn, up to the last three turns are used as the dialogue context. There is a group of external knowledge for each agent turn, which contains multiple knowledge sentences under multiple topics. The bold sentence in each external knowledge group means it is selected for constructing the agent turn.}
\label{d2dexample}
\end{figure*}

Leveraging external knowledge sources besides the dialogue context can help neural conversation models to generate meaningful responses \cite{DBLP:conf/aaai/GhazvininejadBC18,DBLP:conf/ijcai/ZhouYHZXZ18}. The document-grounded dialogue (DGD) \citep{DBLP:conf/emnlp/ZhouPB18,DBLP:conf/emnlp/MogheABK18,DBLP:conf/interspeech/GopalakrishnanH19} uses a document (News report, Wikipedia article, etc) as the external knowledge. A document contains multiple logically related entries (passages or sentences) that together constitute a description of the topic of the document. Figure \ref{d2dexample} shows an example of DGD in the Wizard-of-Wikipedia dataset \cite{DBLP:conf/iclr/DinanRSFAW19}. The dialogue agent is required to construct the responses with knowledge entry in the document. 

Two main challenges in the DGD task are \textbf{knowledge selection} (KS) \cite{DBLP:conf/iclr/DinanRSFAW19,DBLP:conf/iclr/KimAK20,DBLP:conf/emnlp/WuLHO21} and \textbf{response generation} (RG) \cite{DBLP:conf/acl/RashkinRT020}. KS is to select context-relevant knowledge entries from the document. RG is to use dialogue context and the selected information to generate a response. Both KS and RG rely on the correct understanding of the dialogue context and they are closely related to each other through the selected knowledge entry. Current work usually adopted different varieties of encoders (vanilla Transformer \cite{DBLP:conf/nips/VaswaniSPUJGKP17} or Pre-trained language models \cite{DBLP:conf/naacl/DevlinCLT19,radford2019language,DBLP:conf/acl/LewisLGGMLSZ20}) to obtain the semantic representations, then used implicit interaction results between the context and external knowledge for KS and RG \cite{DBLP:conf/aaai/MengRCM0R20,DBLP:conf/acl/QinGBLGDCG19,DBLP:conf/acl/LiNMFLZ19,DBLP:conf/iclr/KimAK20,DBLP:conf/naacl/PrabhumoyeHZBS21}. \textit{However, these methods could not provide an interpretable way to reflect whether the knowledge utilization is based on the correct understanding of the dialogue context}, i.e. \textbf{under what dialogue purpose a knowledge entry is selected and whether it is properly used}.

To improve the interpretability of DGD models, some researchers tried to utilize explicit signals to assist the dialogue modeling. \citet{DBLP:conf/emnlp/FengWGPJL20} showed that human-annotated, task-specific dialogue acts could help the dialogue modeling and the KS in DGD. \citet{DBLP:conf/naacl/ZhanZCDBL21} modeled the knowledge transition in dialogue context with topic tags and leveraged the LDA model to obtain topic tags when they were not available in the dataset. \citet{DBLP:conf/inlg/HedayatniaGKLEH20} used generic, model-annotated dialogue acts, along with topics and knowledge sentences, to form a generation policy to plan the content and style of the response. \citet{DBLP:conf/acl/RashkinRT020} added control tokens that quantified the informativeness and objectivity of response in the training process. These control tokens could help the response to be faithful to selected knowledge and have a required talking style. Similarly, \citet{DBLP:conf/aaai/HazarikaNH22} learned control codes from sets of phrases and then used these codes to generate responses with desired attributes; \citet{DBLP:conf/acl-convai/SahaDS22} used Big-5 personality traits and discourse intent as stylistic control signals for desired response style. \textit{However, previous works used explicit signals only for KS \cite{DBLP:conf/emnlp/FengWGPJL20} or RG \cite{DBLP:conf/interspeech/GopalakrishnanH19,DBLP:conf/aaai/HazarikaNH22}}\footnote{\citet{DBLP:conf/acl-convai/SahaDS22} performed KS in training but not in testing, their method could not show the KS accuracy.}, \textbf{they isolated the connection between KS and RG and could not show whether their signals could provide global guidance for the DGD task.} 

In this paper, we propose an explicit dialogue policy to assist the dialogue modeling in both KS and RG. The dialogue policy is obtained by analyzing the dialogue mode of the DGD task. As shown in Figure \ref{d2dexample}, the conversations from each speaker own their communication intents and these intents are closely related to knowledge selection and injection. For instance, the user starts the conversation with a "Physical fitness" topic and a question. The agent's turn (1) selects a related entry under the "Physical fitness" topic and constructs the response with the selected information. The second user turn refers to the "right kind of physical fitness plan", it leads the agent's turn (2) to a new topic "Fitness culture" which covers different kinds of physical exercise plans. The following user turn centers on this "Fitness culture" topic and introduces more personal opinions. To maintain the conversation, turn (3) of the agent follows the personal information of the user and asks a topic-related question with a knowledge entry under the "Fitness culture" topic.

The dialogue in Figure \ref{d2dexample} shows two kinds of communication intents. \textbf{The first} is the utterance function, e.g. seeking or providing information, assertions, promises, and suggestions. The utterance function is related to the style of an utterance and reflects the dialogue's purposes. \textbf{The second} is the topic transfer intent, e.g. following the current topic or changing to a new related topic. The topic transfer intent guides the utilization of topic-related knowledge, whether similar to context or different from context. The utterance function and topic transfer intent work together to guide the dialogue's purposes and knowledge utilization in the DGD task. In this paper, we aim to learn these two guiding signals as our dialogue policy and design a policy-driven DGD model that can effectively utilize these guiding signals. In this way, we not only leverage the semantic information in these signals to improve the dialogue modeling but also provide explicit guidance for the behavior of the model. Our contributions are as follows:

\begin{itemize}
\item To the best of our knowledge, we are the first to propose an explicit dialogue policy to assist both the knowledge selection and response generation of document-grounded dialogue. Our dialogue policy contains two utterance-level guiding signals: utterance function and topic transfer intent, they are closely related to the dialogue modeling and knowledge utilization in the DGD task.
\item To effectively incorporate the dialogue policy into dialogue modeling, we design a policy-driven document-grounded dialogue (PD-DGD) framework. The PD-DGD adopts different mechanisms in KS and RG to utilize dialogue policy. In this way, we enhance the DGD model with an interpretable way to show under what dialogue policy a knowledge entry is selected and used. 
\item Experiments on three public DGD datasets\footnote{We choose three public DGD benchmarks: WoW \cite{DBLP:conf/iclr/DinanRSFAW19}, Holl-E \cite{DBLP:conf/emnlp/MogheABK18} and Doc2Dial \citep{DBLP:conf/emnlp/FengWGPJL20}. They all have ground-truth knowledge labels that can be used to test the KS accuracy. Details about the datasets are introduced in Section \ref{data}.} show that PD-DGD achieves state-of-the-art performance on KS and RG. We give a detailed analysis of the experiments including a comparison with dialogue policies used by previous researchers. Our code and data will be released on github.com\footnote{https://github.com/malongxuan/PD-DGD}.
\end{itemize}

\section{Related Work}
Knowledge selection (KS) \cite{DBLP:conf/iclr/DinanRSFAW19,DBLP:conf/aaai/RenCM0R20,DBLP:conf/sigir/MengRCSRTR20} and response generation (RG) \cite{DBLP:conf/acl/LiNMFLZ19,DBLP:journals/corr/abs-1911-09728,DBLP:conf/naacl/PrabhumoyeHZBS21} are the two main sub-tasks in DGD. KS is to select context-relevant knowledge information (in the form of words, fragments, or sentences) from the document. RG is to generate a response using the dialogue context and the selected information. In this paper, we categorize the DGD models into two classes. \textbf{Models in the first class} only take the dialogue context and the knowledge entries as input. They do not leverage other explicit signals (e.g. the dialogue act of each utterance) for the KS and RG tasks. Thereby, the performance of DGD models is only based on the interaction results between context and knowledge. For example, \citet{DBLP:conf/iclr/KimAK20} and \citet{DBLP:conf/sigir/MengRCSRTR20} traced the knowledge utilization with each turn of the context and leveraged these clues for the knowledge selection of the response turn; \citet{DBLP:conf/emnlp/ZhaoWXTZY20} jointly modeled the KS and RG with Reinforcement Learning; \citet{DBLP:conf/emnlp/ZhanSCZ21} modeled KS and RG in a Collaborative Latent space. \textbf{Models in the second class} leverage other explicit signals (e.g. signals reflect the intent or styles of the speakers) for the KS or RG tasks. These explicit signals serve as semantic guidance to assist the performance of DGD models. For instance, \citet{DBLP:conf/emnlp/FengWGPJL20} showed that concatenating dialogue context with human-annotated, domain-specific dialogue act labels was useful for KS; \citet{DBLP:conf/inlg/HedayatniaGKLEH20} combined model-annotated dialogue acts, topics, and knowledge sentences as generation policy to control the content and style of the generation; some researchers \cite{DBLP:conf/acl/RashkinRT020,DBLP:conf/acl-convai/SahaDS22,DBLP:conf/aaai/HazarikaNH22} added control tokens (either predefined or learned) to the input sequence of the generative models for injecting certain semantic features. One of the advantages of using external signals is that they can provide interpretation about the behavior of the DGD model, i.e. under what dialogue intent a knowledge entry is selected \cite{DBLP:conf/emnlp/FengWGPJL20} or under what generation policy a knowledge entry is used \cite{DBLP:conf/acl/RashkinRT020,DBLP:conf/aaai/HazarikaNH22}. However, the external signals introduced by previous work were usually designed for KS or RG only and could not provide global interpretation for DGD tasks. \textit{Different from previous work, we propose a dialogue policy to assist the dialogue modeling in both KS and RG, showing under what dialogue policy a knowledge entry is selected and used, adding a global interpretation for the DGD model.}

It is worth noting that the dialogue policy we used in DGD is different from the policies in task-oriented dialogue \cite{DBLP:conf/aaai/HeDZWCLJYHSSL22,DBLP:conf/coling/HeDHYCDHSL22,DBLP:conf/sigir/HeDYSHSL22} since the dialogue policy in the latter is more finely defined and more closely related to response than in DGD. For instance, an "inform" dialogue act in task-orientated dialogue is usually followed with slot types such as "name/location/price" and the corresponding response will provide the related information \cite{DBLP:conf/acl/ChenCQYW19,DBLP:conf/acl/WangTWQY20}. In contrast, the "inform" in DGD is a vague description and the corresponding response can be arbitrary information. Our dialogue policy is also different from open-domain dialogue research without external knowledge \cite{DBLP:conf/naacl/FangCSCHCSO18,DBLP:conf/emnlp/YuCYCWZZJCBISDB19} since they do not need to consider selecting knowledge from an external source.

\begin{table*}[t]
\caption{Interpretation and dialogue examples of the four DA labels.}
\label{DAexample}
\footnotesize
\begin{center} 
\begin{tabular}{l |l}
\hline
\textbf{DA label} & \textbf{Interpretation} \& \textbf{Example}\\
\hline
Question & The speaker wants to know something and seek some information. E.g. "Where is your company?" \\
\hline
Inform & All statements and questions that the speaker provides information. \\
 &E.g. "We are off Singing Road, close to the bank."\\
\hline
Directive & Request, instruct, suggest and accept/reject offer. E.g. "How about getting some coffee for tonight?"\\
\hline
Commissive & Accept/reject request or suggestion and offer. E.g. "I don’t honestly like that kind of stuff."\\
\hline
\end{tabular}
\vspace{-0.5cm}
\end{center} 
\end{table*}

\section{Our proposed Dialogue Policy}
Our dialogue policy consists of two kinds of explicit guiding signals: utterance function and topic transfer intent. In this section, we introduce how to automatically obtain these explicit guiding signals for the utterances in DGD datasets since no human-annotated labels are available.

\begin{table*}[t]
\caption{Dataset statistics with human-annotated DAs for training and testing the DA tagger. "Length/U." means the average length per utterance. The testing data are manually annotated by us.}
\label{trainDAdata}
\footnotesize
\begin{center} 
\begin{tabular}{l |c|c|c|c|c|c}
\hline
\textbf{Dataset} & \textbf{'inform'} DA&  \textbf{'question'} DA& \textbf{'directive'} DA & \textbf{'commissive'} DA& \textbf{Total} & \textbf{Length/U.} \\
\hline
DailyDialog    &  45,469 &   28,994 & 17,267 &  9,296  & 101,026 & 13.8  \\
Switchboard   &  82,176 &   9,097  &  708      &   99      & 92,080    & 10.6  \\
AMI 			&  7,231   &   0          & 3,187   &  549     & 10,967    & 16.2  \\
Maptask          &1,530     &   994      & 2,121   &   0         & 4,645      & 10.7 \\
\hline
Total-for-training & 136,406 & 39,085 & 23,283 & 9,944 &208,718 & 12.5  \\
\hline
WoW-for-testing & 176 & 15   & 8 & 1 & 200 & 16.8   \\
Holl-E-for-testing &  82  & 16  & 2 & 0 & 100 & 15.4 \\
\hline
Total-for-testing & 258 & 31 & 10 & 1 & 300 & 16.3  \\
\hline
\end{tabular}
\end{center} 
\end{table*}

\subsection{Utterance Function}
\label{DAtagger}
We choose to use an ISO standard Dialogue Act \cite{DBLP:conf/lrec/BuntACCFHLPPRST10,DBLP:conf/lrec/BuntPGPFKP20} as the utterance function because it is defined as "the intention or the function of an utterance in dialogues" \cite{DBLP:conf/inlg/KawanoYN19} and this standard has been applied in many studies of dialogue systems \cite{DBLP:conf/ijcnlp/LiSSLCN17,DBLP:conf/sigdial/SankarR19,DBLP:conf/inlg/HedayatniaGKLEH20}. Without loss of generality, we choose a taxonomy with four DA labels in this ISO standard used by previous work \cite{DBLP:conf/ijcnlp/LiSSLCN17}, which is "inform/question/directive/commissive". These four DA labels represent the communicative functions in human dialogue. Table \ref{DAexample} shows the interpretation and dialogue examples of these DA labels. We choose four commonly used dialogue datasets (DailyDialog \cite{DBLP:conf/ijcnlp/LiSSLCN17}, Switchboard \cite{DBLP:conf/icassp/GodfreyHM92}, AMI \cite{carletta2005ami}, and Maptask \cite{anderson1991hcrc}) to train a DA tagger that could annotate DA labels for each utterance in our experiments (Table \ref{trainDAdata}). These four datasets contain real human conversations from different domains and the utterances in them are manually annotated with ISO standard DA labels. Following \citet{DBLP:conf/coling/MezzaCSTR18}, we map utterances with a lower-level DA to the four higher-level DA labels we choose. For example, utterances with lower-level labels 'choice question'/'check question' are mapped to a higher-level label 'question', and utterances with lower-level labels 'suggest'/'request' are mapped to a higher-level label 'directive'. We process the data based on the code released by \citet{DBLP:conf/coling/MezzaCSTR18}. By keeping the utterances longer than 2 words\footnote{For instance, there are 223,607 utterances in Switchboard, we only keep 92,080 of them.}, we get 208,718 utterances from DailyDialog/Switchboard/AMI/Maptask, the label distribution is \{'inform': 136,406, 'question': 39,085, 'directive': 23,283, 'commissive': 9,944\}, the average utterance length is 12.5. We randomly split these data into train and validation (19:1) and train a DA tagger to learn the common dialogue patterns in these utterances. 

We choose BERT \cite{DBLP:conf/naacl/DevlinCLT19} as the DA tagger since it has been proven to be efficient in classification tasks. The input to the DA tagger is an utterance and the output is the corresponding DA label for this utterance. We add an MLP to the output vector of BERT's first input position (a special "<cls>" token) to predict the DA label. After training, we test the capability of the DA tagger with a manually annotated test set. We randomly select 300 utterances from the DGD datasets (200 utterances from WoW \cite{DBLP:conf/iclr/DinanRSFAW19} and 100 utterances from Holl-E \cite{DBLP:conf/emnlp/MogheABK18}) and then manually annotate them with "inform/question/directive/commissive" labels. Table \ref{trainDAdata} presents the statistic of the test set. We use the trained DA tagger to automatically annotate the 300 utterances and compare the results with the manually annotated labels. The DA tagger achieves 88.4\%/100\%/100\%/100\%/90.3\% accuracy for "inform"/"question"/"directive"/"commissive"/"total" of the testing data, respectively. Since some labels only consist of very limited examples in this experiment. We conduct another experiment to verify the performance of the DA tagger. We randomly select 200 examples (50 per DA label) from the model-annotated data shown in Table \ref{DA}. We recruited 3 students who majored in English to conduct manual annotation for these examples. They were asked to give one of the 4 DA labels for the 200 examples. The definition and examples of the DA labels such as in Table \ref{DAexample} were presented to the annotators so that they were aware of the rule. When determining the final result, the majority was adopted when there was a disagreement among the three students. There were a few cases where the three students disagree with each other, we determined these cases by ourselves. We consider the manual annotation as ground truth and report the \textbf{Precision/Recall/F1} for model-annotated results. For each DA label, we define the 50 examples of this DA label as positive and the rest 150 examples of the other DA labels as negative. The results are shown in Table \ref{recognition}. We can see that the "inform" DA has a lower score than others, especially on Precision. It means the DA tagger classified some wrong utterances into the "inform" category. The "directive" has 100\% precision and the highest Recall/F1, the "question" and "commissive" have a very close performance to "directive". This experiment shows that the DA tagger can identify DAs very well.

\begin{table*}[t]
\footnotesize
\centering
\caption{DA label prediction results.}
\label{recognition}
\begin{tabular}{ c | c  c  c  | c | c c c}
\hline
\textbf{DA label} & \textbf{Precision} & \textbf{Recall} & \textbf{F1} 
& \textbf{DA label} & \textbf{Precision} & \textbf{Recall} & \textbf{F1} \\
\hline
"inform"                & 0.9200      & 0.9583      & 0.9388  & 
"question"             & 0.9800      & 0.9608      & 0.9703  \\
"directive"             & 1.0000      & 0.9804      & 0.9901  &
"commissive"       & 0.9800      & 0.9800      &  0.9800  \\
\hline
\end{tabular}
\end{table*}

We also test the tagger on Doc2Dial. There are human-annotated, task-specific dialogue act labels in Doc2Dial: four different DAs for the agent ("query condition/respond solution/respond solution positive/respond solution negative") and $4$ different DAs for the user ("query condition/query solution/response negative/response positive"). We select the 18,748 utterances with "query conditions" DA for testing. The DA tagger assigns the "question" label to 90.9\% of them. These testing results show that the DA tagger is well-trained and can be used for our experiments. After annotation, the data statistics are shown in Table \ref{DA}. Notice that the dialogue mode in DGD data with ground-truth knowledge labels is mainly to consult and provide information and it is different from the daily life conversation we used to train the DA tagger. Hence, it is reasonable that "question"/"inform" labels account for the vast majority. We choose to keep the long tail labels "directive"/"commissive" so that our method can be applied to more scenarios. For example, DGD tasks that are without ground-truth knowledge labels will be more similar to the daily life conversation since it is not required to provide information, we leave the experiments for future work. 

\begin{table}[t]
\caption{Statistics of the utterance functions and topic intents for utterances. DA is short for dialogue act. We also present the human-annotated DA of Doc2Dial data for further experiments.}
\label{DA}
\footnotesize
\centering
\begin{tabular}{c|c|c}
\hline
\textbf{WoW / Holl-E / Doc2Dial} & \textbf{WoW / Holl-E / Doc2Dial} & \textbf{Doc2Dial}\\ 
\hline
\textbf{Model annotated DA} & \textbf{Model annotated Topic transfer intent} & \textbf{Human annotated DA} \\
\hline
inform (159.6K / 76.8K / 20.8K) & mining the initial topic  (137.2K / 24.8K / 8.9K)& query condition (6.8K)\\
question (41.7K / 14.3K / 5.3K) & starting a new topic  (35.6K / 37.4K / 10.1K )& respond solution (18.6K) \\
directive (664 / 320 / 167) &following a new topic  (28.8K / 29.0K / 7.3K) & respond solution pos (511)\\
commissive (39 / 10 / 18)  &                                                                                    & respond solution neg (407)\\
\hline
\end{tabular}
\end{table}

\subsection{Topic Transfer Intent}
\label{Topic-classifier}
Besides the DA information, we also want to annotate the topic intent of an utterance, i.e. whether an utterance is maintaining the current topic or starting a different topic. Previous work used topic models, such as Latent Dirichlet Allocation (LDA) \cite{DBLP:conf/emnlp/BahetiRLD18,DBLP:conf/naacl/ZhanZCDBL21} or trained topic classifier \cite{DBLP:conf/sigdial/YuXBR16,DBLP:conf/naacl/GhazarianLCMGP21}, to extract the topic information in utterances. However, these methods did not directly model the speaker's intent. In this paper, we propose to use a direct topic intent signal which is closely related to the knowledge utilization of the DGD task. 

We noticed that the DGD datasets (WoW, Holl-E, and Doc2Dial) we used all have latent topic transfer labels that can be extracted with a simple rule-based method. For example, in Figure \ref{d2dexample}, the external knowledge for each utterance is entries collected with several topical phrases. The conversation has an initial topic phrase "Physical fitness". The first turn of dialogue is always "mining the initial topic". When the conversation continues with this initial topic, the dialogue intent can be denoted by "mining the initial topic", such as the second and third turns in Figure \ref{d2dexample}. When one speaker starts a new topic different from the initial topic, the dialogue intent can be denoted by "starting a new topic", such as the fourth turn (the second agent turn) in Figure \ref{d2dexample} which changes the topic to "Fitness culture". When the other speaker follows the new topic, the dialogue intent for the second person is "following a new topic", such as the fifth~seventh turn in Figure \ref{d2dexample}. If the conversation topic changes from "the new topic" to the initial topic "Physical fitness", the dialogue intent can still be denoted by "mining the initial topic". Similarly, in the Holl-E dataset, the external knowledge for each utterance is four groups of entries (plots, reviews, comments, fact table) about a movie. The topic in the conversation is changing among the four knowledge sources. We can also annotate "mining the initial topic", "starting a new topic", and "following a new topic" for each utterance of the Holl-E data according to the topic transfer among different knowledge sources. Similar to WoW and Holl-E, external knowledge in Doc2dial is a document consisting of several paragraphs that focus on different topics. Each agent utterance is grounded on a span of the paragraph. We can extract the same topic intent as WoW/Holl-E for each utterance according to the topic transfer among different paragraphs.

Using the rules described above, we obtain the "mining the initial topic"/"starting a new topic"/ "following a new topic" labels in Table \ref{DA}. For each dataset, we train a topic intent classifier with the annotated train set. The classifiers are also BERT-base models and trained similarly to the DA tagger. The topic intent classifiers take the "dialogue context, the topic intent of the context, and response" as input. The output is the topic intent label of the response. After training, we use these intent classifiers to annotate the test set of WoW, Holl-E, and Doc2Dial. We compare the annotated labels with the ground-truth rule-based labels on the test sets, the accuracy of the annotating is 91.6\%. Next, we will introduce how to use the model-annotated DA labels and topic-intent labels for our PD-DGD model.

\section{Our Proposed Model}

\subsection{Problem Statement}
Given a set of knowledge collections \textbf{K} = [$K_1$,$K_2$,...,$K_{|\textbf{K}|}$] with $|\textbf{K}|$ sentences as external knowledge, a dialogue context \textbf{C} = [$C_1$,$C_2$,...,$C_{|\textbf{C}|}$] with $|\textbf{C}|$ turns and the response $R$ = [$r_1$,$r_2$,...,$r_{|R|}$] with $|R|$ tokens, the DGD models learns to generate $R$ with probability $P$($R$$\mid$$\textbf{K},\textbf{C};\Theta)$, $\Theta$ is the model's parameters. 

After adding utterance function (DA) information $\textbf{D}$ = [$D_1$,$D_2$,...,$D_{|\textbf{C}|}$] and topic transfer intent information $\textbf{T}$ = [$T_1$,$T_2$,...,$T_{|\textbf{C}|}$] for $\textbf{C}$ in this probability, the generation model changes to $P$($R$$\mid$\textbf{K},\textbf{C},\textbf{D},\textbf{T};$\Theta$). When the grounding knowledge is $K_i$, dialogue policy for $R$ are [$D_R$, $T_R$], we can further separate $P$($R$$\mid$\textbf{K},\textbf{C},\textbf{D},\textbf{T};$\Theta$) into \textbf{policy planner} $P_{KS}$($K_i$,$D_R$,$T_R$$\mid$\textbf{K},\textbf{C},\textbf{D},\textbf{T};$\Theta_{KS}$) and \textbf{generator} $P_{RG}$($R$$\mid$$K_i$,\textbf{C},$D_R$,$T_R$; $\Theta_{RG}$), $\Theta_{KS}$ and $\Theta_{RG}$ are models' parameters. The policy planner predicts the response policy while selecting knowledge, then the predicted policy and the selected knowledge are sent to the generator to assist response generation.

\begin{figure}[t]
\centering
\includegraphics[width=\linewidth]{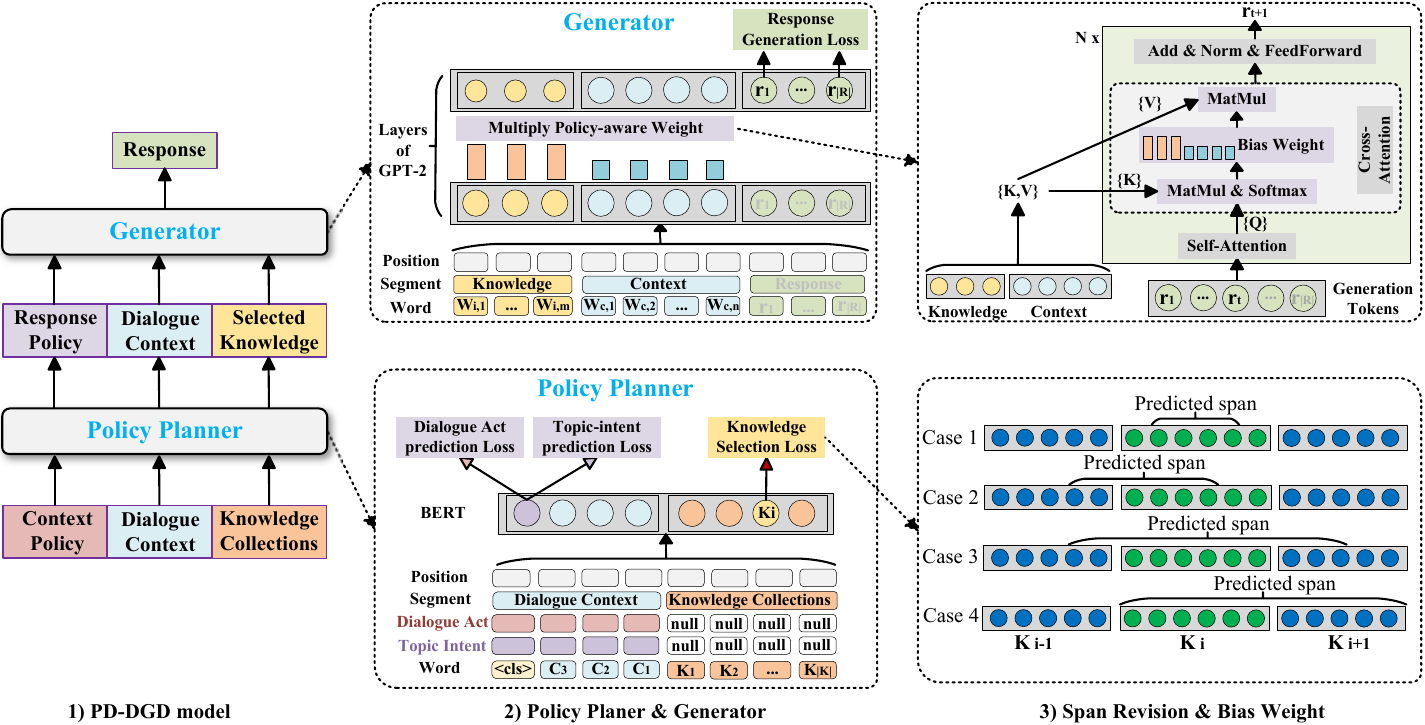}
\caption{The architecture of the PD-DGD model. Notice that $C_i$ and $K_i$ are sentences with multiple tokens.}
\label{bertgpt}
\end{figure}

\subsection{Model Structure}
The structure of PD-DGD is shown in Figure \ref{bertgpt} where $P_{KS}$ and $P_{RG}$ are defined with BERT \cite{DBLP:conf/naacl/DevlinCLT19} and GPT-2 \cite{radford2019language}, respectively. 

\subsubsection{The Policy Planner}
We set the dialogue context to the last three turns, and the input to the policy planner is a concatenated sequence [$<$cls$>$;$C_3$;$C_2$;$C_1$;$<$esp$>$;\textbf{K}], where [;] is the concatenation operation; $<$cls$>$ is a special token of the first input position; $<$esp$>$ is a special token between context and knowledge. We associate each of the policy signals (dialogue acts and topic transfer intents) with a specific trainable embedding that is of the same dimension as the input token embedding. These specific embeddings are randomly initialized and can be learned through training. During training, each utterance is assigned a DA label and a Topic intent label. The corresponding DA embedding and Topic intent embedding are added to each token of the utterance when this utterance is input to the model. Hence each token of an utterance is initialized with the sum of \textbf{five} embeddings: \textit{Word}/\textit{dialogue act}/\textit{topic transfer intent}/\textit{Positional} \cite{DBLP:conf/nips/VaswaniSPUJGKP17}/\textit{Segment} \cite{DBLP:conf/naacl/DevlinCLT19}. After training, the trained DA/Topic intent embeddings are fixed and used for testing. Words in \textbf{K} are similarly initialized except without the \textit{DA}/\textit{topic intent} embeddings. The multi-layer bidirectional attention mechanism in BERT allows the dialogue context \textbf{C} and the policy information \textbf{D}/\textbf{T} to sufficiently interact with each other, resulting in policy-aware context representations, which is then used for policy prediction and KS.

For KS, we train the model to predict the start and end positions of a text span. The KS Loss is a Cross-Entropy (CE) loss: 

\vspace{-0.5cm}
\begin{align}
\mathcal{L}_{KS} =&  - \frac{1}{N} \sum_{i=1}^{N} ( logP(y_i^s) + logP(y_i^e) ),
\end{align}

where $N$ is the number of training samples, $y_i^s$ and $y_i^e$ are the ground-truth start and end positions of the knowledge sentence, respectively. We use a \textit{Span Revision} method that forces the model to predict a whole sentence instead of random positions. As shown in "3) Span Revision \& Bias Weight" in Figure \ref{bertgpt}, when the start and end positions are within or across sentences, we expand, move, or truncate a span into a whole sentence. All 4 cases in Figure \ref{bertgpt} select $K_i$ as the knowledge after revision. Compared with the previous methods \cite{DBLP:conf/iclr/KimAK20,DBLP:conf/sigir/MengRCSRTR20} of encoding candidate sentences into a vector representation, our method can better use the semantic information between sentences. The reasons include: 1) The multi-head attention mechanism in BERT provides sufficient interaction between dialogue and knowledge sentences at the word level, so as to leverage the overall document information for selection; 2) The word-level interaction is consistent with the pre-training process of BERT, so as to fully leverage the ability of the pre-trained model. Another advantage of the span revision method is that it can be applied to other scenarios. The DGD tasks can be categorized into two classes: with or without ground-truth knowledge labels. Meanwhile, the knowledge used by dialogue can be in different formats (phrases, a text span across sentences, paragraphs, multiple sentences, etc.). Our span revision methods can be easily adapted to different DGD tasks (with or without ground-truth knowledge labels, and different knowledge formats).

For the response policy, we pass the last BERT layer's representation of the special token $<$cls$>$ into multi-layer perceptions (MLP) to predict $D_R$ and $T_R$, respectively. The DA/topic-intent prediction losses $\mathcal{L}_{DA}$/$\mathcal{L}_{Topic}$ are CE losses between predicted labels $y_i^D$/$y_i^T$ and ground-truth DA label $\bar{y}_i^D$/$\bar{y}_i^T$: 

\vspace{-0.5cm}
\begin{align}
\mathcal{L}_{DA} =  - \frac{1}{N} \sum_{i=1}^{N} (\bar{y}_i^D logP(y_i^D) ), \ \ \ \ \ \ \ \ \mathcal{L}_{Topic} =  - \frac{1}{N} \sum_{i=1}^{N} (\bar{y}_i^T logP(y_i^T) ).
\end{align}

During training, the KS module needs to select the accurate knowledge sentence and predict the correct response policy simultaneously. The $\mathcal{L}_{DA}$ and $\mathcal{L}_{Topic}$ are easier to be trained since the DA and topic-intent only have a few categories. We use the uncertainty loss \cite{DBLP:conf/cvpr/KendallGC18} to balance the three objectives: 

\vspace{-0.5cm}
\begin{align}
\mathcal{L}_{PolicyPlanner} =& \frac{1}{\mu_1^2} (\mathcal{L}_{DA} + \mathcal{L}_{Topic})+ \frac{1}{\mu_2^2} \mathcal{L}_{KS} + log(\mu_1\mu_2),
\end{align}

where $\mu_1$ and $\mu_2$ are parameters to be learned.

\begin{table*}[t]
\caption{Bias weight vector we used in the generator. The vector has an equal length with the input sequence to the generator (i.e. the selected knowledge entry $K_i$ and the dialogue context \textbf{C}). For example, the bias weight vector will be m+n tokens if $K_i$ has m tokens and \textbf{C} has n tokens. These weights are determined empirically.}
\label{bias-examples}
\footnotesize
\center
\begin{tabular}{c|c|c}
\hline
\textbf{utterance function (DA)} & \textbf{topic transfer intent} & \textbf{Bias weight vector (m+n)}\\
\hline
inform            & starting a new topic &  [$4_1$,$4_2$,...,$4_m$,$1_1$,$1_2$,...,$1_n$] \\
inform            & mining the initial topic & [$3_1$,$3_2$,...,$3_m$,$1_1$,$1_2$,...,$1_n$]\\
inform            & following a new topic & [$2_1$,$2_2$,...,$2_m$,$1_1$,$1_2$,...,$1_n$]\\
\hline
question / directive / commissive   & starting a new topic & [$3_1$,$3_2$,...,$3_m$,$1_1$,$1_2$,...,$1_n$]\ \\
question / directive / commissive   & mining the initial topic & [$2_1$,$2_2$,...,$2_m$,$1_1$,$1_2$,...,$1_n$]\\
question / directive / commissive   & following a new topic & [$1_1$,$1_2$,...,$1_m$,$1_1$,$1_2$,...,$1_n$]\\
\hline
\end{tabular}
\end{table*}

\subsubsection{The Generator}
\label{thegenerator}
After the knowledge selection, we plan to incorporate the dialogue policy information into the response generation process. We observe that comparing the "inform" DA with "question"/"directive"/"commissive", the former is expected to generate a response using more external knowledge information than the context because the "inform" DA usually means to introduce new information from external knowledge. The other DAs are not required to introduce external knowledge as the "inform" DA and they mainly focus on the information already in the context. For example, we usually ask questions based on the information already mentioned in the context, we do not need to introduce external knowledge. Similarly, comparing the "starting a new topic" with the "following a new topic" intent, the former is required to introduce new information from the external knowledge and the latter can use the topic information already in the context. Hence, we introduce extra weight for the selected knowledge when the predicted dialogue policy includes dialogue policies such as "inform" and "starting a new topic". This is done by directly adjusting the attention distribution in the cross-attention layer of the generator \cite{DBLP:conf/aaai/HazarikaNH22}.

As shown in Figure \ref{bertgpt}, the input to the Generator is [$W_{i,1}$, ..., $W_{i,m}$, $W_{c,1}$, ..., $W_{c,n}$], where $W_{i,m}$ represents the $m$-th token in selected knowledge entry $K_i$ and $W_{c,n}$ represents the $n$-th token in dialogue context \textbf{C}. Tokens are initialized with the sum of Word/Segment/Positional embeddings. Each Transformer layer of the Generator contains a self-attention and a cross-attention, we add a bias weight to $K_i$ in the cross-attention layer. More formally, we denote the encoding result of [$W_{i,1}$, ..., $W_{i,m}$, $W_{c,1}$, ..., $W_{c,n}$] as $\textbf{e}_{K_i, \textbf{C}}$, attention matrices in Transformer layer are $W_Q$, $W_K$ and $W_V$ $\in$ $\mathbb{R}^{d*d}$, and the output of the self-attention layer $\textbf{e}_t$ $\in$ $\mathbb{R}^{d}$ for $r_t$ (the $t$-th token of $R$), the cross-attention output for $r_t$ is :

\vspace{-0.5cm}
\begin{align}
\mathcal{N} \left( \textbf{b} \odot \text{Softmax}\left( \frac{(\textbf{e}_t W_Q) (\textbf{e}_{K_i, \textbf{C}} W_K)^T}{\sqrt{d}} \right) \right) \textbf{e}_{K_i, \textbf{C}}W_V,
\end{align}

where $\mathcal{N}$ is normalization, $\odot$ is element-wise multiplication and \textbf{b} is the bias weight vector. We show examples of \textbf{b} in Tabel \ref{bias-examples}. A more direct way is to automatically learn the bias weight with the input information (i.e. the dialogue context and the policy for the response). However, we test with several settings and the results are not as good as the empirical weight. The experiments and analysis are shown in Section \ref{automaticweight}. We leave the automatic learning of the bias weight vector for future work. \textbf{b} works through element-wise multiplication with the attention distribution and then normalizing the results so that the outcome is still a probability distribution. In an auto-regressive model such as GPT-2, context first interacts with the policy-aware $K_i$ to get policy/knowledge-aware context representations, then the interaction results are used for generation. The RG loss is:

\vspace{-0.5cm}
\begin{align}
\mathcal{L}_{Generator} =&  - \frac{1}{N} \sum_{i=1}^{N}\sum_{t=1}^{r} ( logP(R_i^t) ),
\end{align}

where $R_i^t$ is the t-th word of the i-th response. The policy planner and generator are fine-tuned separately and we adopt a joint training step to combine them into a DGD framework. 

\subsubsection{Joint Training}
Following \citet{DBLP:conf/emnlp/ZhaoWXTZY20}, we adopt a Reinforcement Learning procedure to jointly train the policy planner and generator. We use the policy-gradient method \cite{DBLP:conf/nips/SuttonMSM99} to continue-train the policy prediction in the KS module. Specifically, the $\mathcal{L}_{DA}$ and $\mathcal{L}_{Topic}$ in equations (2) and (3) are further "supervised" by the generator and are directly optimized with F1 metric:

\vspace{-0.5cm}
\begin{align}
\mathcal{L}_{RL} =&  - \frac{1}{N} \sum_{i=1}^{N}(F_i^1(\bar{y}_i^D logP(y_i^D) + \bar{y}_i^T logP(y_i^T) ),\\
F_i^1 =& F1(\bar{R}_i, R_i)- \frac{1}{N}\sum_{j=1}^{N}F1(\bar{R}_j, R_j),
\end{align}

where $F1$ computes the F1 metric between generated response $\bar{R}_i$ and the ground truth response $R_i$. Minimizing $\mathcal{L}_{RL}$ aligns the policy prediction results with the generation quality. 

\begin{table*}[t]
\caption{Statistics of Datasets. "W./T./C.K.S." is "Words/Turn/Candidate Knowledge Sentences", respectively.}
\label{statistics}
\footnotesize
\center
\begin{tabular}{l|c|c|c|c|c|c}
\hline
Dataset & Dialogues (Train/Validation/Test)  & T.s/Dialog & W./T. &External Source & W./Source &C.K.S./T. \\
\hline
WoW            & 22,311 (18,430 / 1,948 / 1,933) & 9.1 & 17.2 & 1,356,509 (sentences) & 30.7& 61.2 \\
Holl-E           &  9,071 (7,228 / 930 / 913) &10.0& 15.3 & 921 (documents) & 727.8 & 57.6\\
Doc2Dial      & 4,135 (3,474 / 661 / ---) & 15.6 & 14.0 & 458 (documents) & 947.0 & 73.1 \\
\hline
\end{tabular}
\end{table*}

\section{Experimental Setup}

\subsection{Datasets}
\label{data}
Tabel \ref{statistics} shows the data statistics. Doc2Dial has human-annotated communication intents while WoW and Holl-E do not. After using the DA tagger (section \ref{DAtagger}) and the topic intent classifiers (section \ref{Topic-classifier}) to annotate labels on them, we can compare our model-annotated policy with human-annotated ones on Doc2Dial and verify the effect of using dialogue policy information on all three datasets. We use the $1.0.1$ version of Doc2Dial and use the validation set for testing since we do not have the access to the test set. There are Test seen/unseen sets in WoW according to whether including topics not seen in the training set and Test single/multi-reference sets in Holl-E according to one/multiple ground-truth responses to a dialogue context.

\subsection{Baselines}
We compare our model with two groups of baselines. The first group is models with similar encoder or decoder structures to our model but without a dialogue policy. We use models in this group to compare the performance when introducing our dialogue policy. The second group is models with explicit dialogue policies. As we introduced in section \ref{introduction}, models in this group usually focus on using dialogue policies for generation \cite{DBLP:conf/inlg/HedayatniaGKLEH20,DBLP:conf/acl-convai/SahaDS22,DBLP:conf/aaai/HazarikaNH22}, they did not perform the knowledge selection task so we could not compare the knowledge selection results with them. For these models, we choose to apply their dialogue policy in BERT-base for comparing different approaches.

Models in the first group include: \textbf{(1)} Transformer Memory Network (\textbf{TMN}) \citep{DBLP:conf/iclr/DinanRSFAW19} uses Transformer structure for KS and is introduced along with WoW dataset; \textbf{(2)} Sequential Knowledge Transformer (\textbf{SKT}) \citep{DBLP:conf/iclr/KimAK20} uses BERT as encoder and selects knowledge with a sequential latent variable model; \textbf{(3)} Dual Knowledge Interaction Network (\textbf{DukeNet}) \citep{DBLP:conf/sigir/MengRCSRTR20} uses BERT as encoder and proposes a knowledge shifter and tracker module for KS; \textbf{(4)} \textbf{KnowledGPT} is a state-of-the-art DGD model proposed by \citet{DBLP:conf/emnlp/ZhaoWXTZY20}. It uses reinforcement learning to optimize knowledge selection and response generation\footnote{We use the WoW dataset to compare with KnowledGPT since the authors only provide the evaluation code for WoW.}; \textbf{(5)} Collaborative Latent Variable model (\textbf{CoLV}) \cite{DBLP:conf/emnlp/ZhanSCZ21} integrates knowledge selection and response generation in separate yet collaborative latent spaces. We choose these baselines because 1) they provide KS accuracy results and SKT/ DukeNet / KnowledGPT / CoLV / PD-DGD all employ BERT-base (110M) as encoder, so we can fairly compare the KS accuracy between them; 2) KnowledGPT / PD-DGD both use GPT-2 (117M) as generation module, we can fairly compare the generation quality between them.\footnote{It is worth noting that we only compare works that perform knowledge selection using the original knowledge entries, we did not compare works using more document clues \cite{DBLP:conf/emnlp/WuLHO21,DBLP:conf/coling/XuZFKN22}. For example, a most recent work \cite{DBLP:conf/coling/XuZFKN22} reconstructed the document into a multi-document graph structure. The work incorporated co-referential mentions and reconstructed the document structure, which is a different setting from the baselines we used. Whether our dialogue policy can work together with external clues is our future work.}

Policy in the second group include\footnote{We use the same BERT-base model as in PD-DGD for these policies for a fair comparison.}: \textbf{(1)} human-annotated DA on Doc2dial \cite{DBLP:conf/emnlp/FengWGPJL20} (shown in Table \ref{DA}). Following \citet{DBLP:conf/emnlp/FengWGPJL20}, we concatenate the human-annotated DA descriptions in front of the input to our policy planner to perform knowledge selection and next DA prediction. Then we concatenate the predicted DA, the selected knowledge, and the dialogue context as the input to a GPT-2 model. These two models are denoted by \textbf{BERT/GPT-2(manual policy)}, respectively; \textbf{(2)} a learned control code used by \citet{DBLP:conf/aaai/HazarikaNH22}. The control code is an embedding learned from a set of phrases/sentences. For example, we randomly select a set of sentences with the same DA label "question", then use the encoder layer of PD-DGD to encode these sentences into embeddings. Then the average of these sentence embeddings is used as the control code for the "question" label. In this way, we can have a control code for each guiding signal in our dialogue policy. Following \citet{DBLP:conf/aaai/HazarikaNH22}, we concatenate these control codes in front of the input to our policy planner to perform knowledge selection and the next DA prediction. Then we concatenate the predicted control code, the selected knowledge, and the dialogue context as the input to a GPT-2 model. Notice that the control code for GPT-2 is learned with the encoder of GPT-2 and also learned from a set of sentences with the same label. This setting is denoted by \textbf{BERT/GPT-2(learned policy)}; \textbf{(3)} a stylistic-control code \cite{DBLP:conf/acl-convai/SahaDS22} that includes Big-5 Personality Traits, Corpus-based traits, and Subjective/Objective utterance intents. We use the taggers provided by \citet{DBLP:conf/acl-convai/SahaDS22} to obtain these stylistic-control signals for each utterance of the datasets we used. Following \citet{DBLP:conf/acl-convai/SahaDS22}, we concatenate these signals in front of the input for a BERT-base model to perform KS and the next control code prediction. This setting is denoted by \textbf{BERT/GPT-2(stylistic Policy)}; \textbf{(4)} \textbf{BERT/GPT-2(without policy)}: BERT(without policy) is our policy planner taking \textbf{C} and \textbf{K} as input and selecting a knowledge entry $K_i$ from \textbf{K}. GPT-2(without policy) is a GPT-2 model that takes \textbf{C} and $K_i$ as input and generates a response. 

Besides the above two groups of baselines, we also provide different settings of PD-DGD: 1) \textbf{PD-DGD(-DA)}/\textbf{PD-DGD(-Topic)}/\textbf{PD-DGD(-DA,-Topic)} are settings without part or all of the dialogue policy. They can show the contribution of using DA/topic-intent to our model; 2) \textbf{PD-DGD(-$\mathcal{L}_{DA}$)}/\textbf{PD-DGD(-$\mathcal{L}_{Topic}$)}/\textbf{PD-DGD(-$\mathcal{L}_{DA}$,-$\mathcal{L}_{Topic}$)} are settings without part or all of the policy prediction losses. They can show the effectiveness of multi-task learning; 3) \textbf{PD-DGD(-Joint Learning)} shows the effects of joint training when using our policy; 4) \textbf{PD-DGD(- Span Revision)} is our Policy Planner without the span revision method; 5) \textbf{PD-DGD(Sentence)} forces the policy planner to predict one position (instead of both the start and end positions). The position choices are limited to the ends of sentences (instead of arbitrarily positions). In this way, the model can directly select a sentence without the span revision method, which is similar to sentence-level classification; 6) \textbf{PD-DGD(- Bias Weight)} is our Generator without the bias weight, it is equal to \textbf{PD-DGD(-$\mathcal{L}_{DA}$,-$\mathcal{L}_{Topic}$)}; 7) \textbf{PD-DGD(ground-truth)} is a GPT-2 model using model-annotated dialogue policy, ground-truth knowledge sentences, and dialogue context for a generation. It is the generation upper bound of PD-DGD.

\subsection{Implementation Details} 
The implementations of BERT and GPT-$2$ are based on the public Pytorch implementation\footnote{https://github.com/huggingface/transformers}. The hyper-parameters which are not introduced in this section follow the original implementation in the link. During BERT fine-tuning, we truncated the length of the dialogue context to $60$ tokens and the maximum input length to $512$ tokens. Document longer than $512$ tokens is split into multiple examples and we choose the knowledge entry with the highest probability in these examples as the KS result. The maximum predicted span length is set to $90$ words. In GPT-$2$, the beam-search size is 5. Adam optimizer with $\beta_1$ = $0.9$ and $\beta_2$ = $0.999$ is used for all models. We use a single Tesla v$100$s GPU with $32$gb memory to conduct experiments, the batch size is $4$ for all datasets. The fine-tuning epochs are 3 for the policy planner and 4 for the generator. The DA/Topic intent taggers are trained for 3 epochs. In joint optimization, the batch size is set to 4, and the learning rates for the Policy Planer and the Generator are set as 5e-6 and 5e-5 respectively. 

\subsection{Evaluation Metrics}
We use the following automatic evaluation metrics employed by the baselines. For KS, we use Hits@1 \cite{DBLP:conf/iclr/DinanRSFAW19} to measure the KS and DA prediction accuracy. For response generation, we use perplexity (PPL), unigram F1 \cite{DBLP:conf/iclr/DinanRSFAW19}, BLEU-4 \cite{DBLP:conf/acl/PapineniRWZ02}, ROUGE-L \cite{lin2004rouge}, and Distinct-(1/2) \cite{DBLP:conf/naacl/LiGBGD16}. PPL measures the probability of the model predicting the real response. Unigram F1/BLEU-4 measures the 1/4-gram overlap between the generated response and the reference one, respectively. ROUGE is based on the calculation of the recall rate of the common sub-sequence of generating response and the real one. Distinct measures the diversity of responses by calculating the proportion of distinct n-grams in the total number of n-grams. Lower PPL and higher Hits@1/F1/BLEU/ROUGE/Distinct mean better performance. Notice that the PPL can only be compared among the same model when using the same vocabulary.

For manual evaluation, we recruit $3$ professional researchers who focus on natural language processing (NLP) and dialogue systems (DS) and have published papers at international conferences related to NLP and DS. We randomly select $50$/$50$ dialogue samples from the WoW unseen/Doc2Dial validation sets, respectively. The generated responses to these samples are presented to the annotators accompanied by their corresponding dialogue history (3 turns) and external knowledge. The responses from different models are shuffled so the annotators do not know which model the response is coming from. Following \citet{DBLP:conf/emnlp/ZhaoWXTZY20}, we only provide the ground-truth knowledge sentences and ask the annotators to judge the response quality from three aspects: \textit{Fluency}, \textit{Context Coherence} and \textit{Knowledge Relevance}. The annotators assign a score in \{0:bad, 1:fair, 2:good\} to each response for each aspect. The agreement among the annotators is measured via Fleiss’ kappa \citep{Fleiss1971Measuring}.

\begin{table}[t]
\caption{KS results (Hits@1) on the WoW Test seen/unseen, Holl-E Test single/multi-reference, and Doc2Dial Validation sets. The DukeNet is the base model to do the significant test for our models (* means statistically significant with p$<$0.01). Results with "\#" are obtained with the released code. "---" means the previous work did not report the result. "-- --" means the dataset does not have manual policy labels.} 
\label{ks}
\footnotesize
\begin{center} 
\begin{tabular}{l | c | c | c | c | c}
\hline
\multirow{2}{*}{\diagbox{\textbf{Model}}{\textbf{data}}}  &  \textbf{WoW} & \textbf{WoW} & \textbf{Holl-E}   & \textbf{Holl-E}  & \textbf{Doc2Dial}  \\
& \textbf{Test seen}  & \textbf{Test unseen} &\textbf{Test single} &\textbf{Test multiple} & \textbf{Validation}  \\
\hline
TMN (\citet{DBLP:conf/iclr/DinanRSFAW19}) & 21.6 & 12.1 & 22.7 & 32.2  & 43.1\#   \\
SKT (\citet{DBLP:conf/iclr/KimAK20})               & 26.8 & 18.3 &   29.2 & 38.3  & 49.5\#  \\
DukeNet (\citet{DBLP:conf/sigir/MengRCSRTR20})           & 26.4 & 19.6 &  30.0 & 40.3  & 49.7\#   \\
CoLV (\citet{DBLP:conf/emnlp/ZhanSCZ21})                       & 30.1 & 18.9 &  32.6 & --- & ---    \\
KnowledGPT (\citet{DBLP:conf/emnlp/ZhaoWXTZY20}) & 28.0 & 25.4 &  --- & ---  & ---    \\
\hline
BERT (without policy) &     27.3* & 25.1*   &  31.8* & 41.2*   & 56.0*      \\ 
BERT (manual policy \cite{DBLP:conf/emnlp/FengWGPJL20}) &   -- --    & -- --  &  -- --  & -- --  & 59.4*     \\ 
BERT (learned Policy \cite{DBLP:conf/aaai/HazarikaNH22})  &   28.4* & 25.2*    &   32.0* & 41.5*  & 56.6*    \\
BERT (stylistic Policy \cite{DBLP:conf/acl-convai/SahaDS22})  &   26.8* & 25.0*    &   31.0* & 41.0*  & 56.0*   \\
\hline
PD-DGD                 &\textbf{30.7}* & \textbf{29.7}* & \textbf{36.1}* & \textbf{44.5}* & \textbf{59.6}* \\
\ \ \ \ \ \ (- DA)      &     28.8* & 28.1*   &  35.3* & 44.0*   &57.7*  \\
\ \ \ \ \ \ \ \ \ \ \ \ (-$\mathcal{L}_{DA}$)     &     30.3* & 29.2*   &  35.2* & 43.1*   &57.4*  \\
\ \ \ \ \ \ (- Topic)   &     28.9* & 28.0*   &  33.9* & 42.5*   & 57.9*  \\
\ \ \ \ \ \ \ \ \ \ \ \ (-$\mathcal{L}_{Topic}$) &    30.1* & 29.3*   &  35.5* & 43.0*   &57.2*  \\
\ \ \ \ \ \ (- DA, - Topic)  &     27.0* & 25.2*   &  32.9* & 41.6*   & 56.0*  \\
\ \ \ \ \ \ \ \ \ \ \ \ (- $\mathcal{L}_{DA}$,- $\mathcal{L}_{Topic}$)   
                                                   &   29.9* & 28.0*   &  34.4* & 42.8*   & 57.3*  \\
\ \ \ \ \ \ (- Joint Learning)                 &30.0* & 29.1* & 35.3* & 42.6*& 59.1* \\
\ \ \ \ \ \ (- Span Revision)& 25.2* & 24.9* & 33.1* & 40.4*& 59.0* \\
\ \ \ \ \ \ (Sentence) & 29.0* & 26.8* & 34.0* & 41.7*& 59.1* \\
\hline
\end{tabular}
\end{center} 
\end{table}

\section{Experimental Results and Analysis}
We aim to answer the following questions with our experiments. For the knowledge selection: \textbf{(1)} whether our model outperforms baselines that conduct KS without dialogue policies? (section \ref{no-policy}) \textbf{(2)} how does our dialogue policy compare to policies proposed by other researchers in KS? (section \ref{other-policy}) \textbf{(3)} what are the gain in KS comes from? (section \ref{ks-ablation}) \textbf{(4)} what is the difference when applying the DA policy on different dialogue data in KS? (section \ref{DAnaturalness}) \textbf{(5)} what is the difference when applying the Topic intent policy on different dialogue data in KS? (\ref{documentTopic}) \textbf{(6)} can our dialogue policy work on different or larger pre-trained models in KS? (sections \ref{scaling}) \textbf{(7)} why the utterance function and topic transfer contribute to the knowledge selection process? (sections \ref{whycontributes})

For the response generation: \textbf{(8)} whether our model outperforms baselines that conduct RG without dialogue policies? (section \ref{no-policy-rg}) \textbf{(9)} how does our dialogue policy compare to policies proposed by other researchers in RG? (section \ref{other-policy-rg}) \textbf{(10)} what are the gain in RG comes from? (section \ref{rg-ablation}) \textbf{(11)} what is the difference when applying the DA policy on different dialogue data in RG? (section \ref{DAnaturalnessRG}) \textbf{(12)} what is the difference when applying the Topic intent policy on different dialogue data in RG? (\ref{TopicnaturalRG}) \textbf{(13)} can our dialogue policy work on different or larger pre-trained models in RG? (section \ref{ScalingGen}) \textbf{(14)} what can we learn from manual evaluation and case study? (sections \ref{manual-eval} and \ref{case-study}) \textbf{(15)} why not automatically learn the bias weight? (section \ref{automaticweight}) \textbf{(16)} how do we decide the bias weight? (section \ref{decide-bias-weight})

\subsection{Knowledge Selection (KS)}

\subsubsection{Comparing with baselines without dialogue policy}
\label{no-policy}
Table \ref{ks} shows the KS accuracy results of all models. We can see that SKT/DukeNet have fairly close performance, and they both outperform TMN. This is because SKT/DukeNet employ the same BERT-base encoder, and the representation ability of the BERT encoder helps them better align knowledge entry with dialogue context than the vanilla transformer encoder in TMN. CoLV is better than SKT/DukeNet on WoW seen and Holl-E single Test sets because the collaborative latent variable model can better integrate KS and RG. However, CoLV shows its limitation when applying to WoW Test unseen set. KnowledGPT also leverages BERT-base as an encoder, it uses an LSTM after the BERT to sequentially select knowledge sentences and adopts a weakly supervision signal from the generation module to guide the knowledge selection. Benefiting from its learning policy, KnowledGPT surpasses CoLV by 6.5 on WoW Test unseen. Compared with baselines without dialogue policy, our PD-DGD outperforms DukeNet 3.2 on the Holl-E Test multi-reference and exceeds CoLV 3.5 on the Holl-E Test single-reference. On the Doc2Dial validation set, PD-DGD surpasses DukeNet 9.9. On the WoW Test seen/unseen, PD-DGD outperforms KnowledGPT 2.7/4.3 and surpasses CoLV 0.6/10.8, respectively. When removing our dialogue policy, PD-DGD(-DA,-Topic) model only has 27.0/25.2 KS accuracy on WoW, which is close to KnowledGPT. These results show that PD-DGD successfully integrates the semantic information of our dialogue policy to help the KS process.

\subsubsection{Comparing with other dialogue policies}
\label{other-policy}
The second group of Table \ref{ks} shows the KS accuracy results when using dialogue policies proposed by other researchers. Firstly, we can see that BERT(manual policy) outperforms BERT(without/learned/stylistic Policies) on Doc2Dial. This is reasonable since the manual policy is more accurate to describe the dialogue intention and shares more semantic overlaps with the dialogue. BERT(learned policy) is better than BERT(without policy), this shows that \textbf{1)} our policy planner module is capable of leveraging different kinds of dialogue policies to assist KS; \textbf{2)} learning semantic information from a set of control phrases with similar properties can assist dialogue understanding. On the other hand, BERT(stylistic policy) is slightly worse than BERT(without/learned policy), this result shows that the stylistic policy which aims to improve the generation quality could not help the KS task in a similar setting. Overall, PD-DGD is better than all policies proposed by previous researchers, including the strong manual policy on Doc2Dial. These results show that our dialogue policy can help build the correlations between dialogue and knowledge entries, and consequently guide the knowledge selection.

\subsubsection{Ablation Study in KS}
\label{ks-ablation}
Table \ref{ks} shows the ablation study results of PD-DGD in different settings. We can see that \textbf{1)} PD-DGD is better than PD-DGD(-DA)/(-Topic)/(-DA,-Topic), PD-DGD(-DA)/(-Topic) are better than PD-DGD(-DA,-Topic). These results mean that PD-DGD benefits from both the DA and Topic-intent signals when performing KS; \textbf{2)} PD-DGD(-DA) is better than PD-DGD(-Topic). This shows that removing Topic-intent causes more declines than removing DA in our policy. This is because the topic intent is more closely related to the knowledge selection since it reflects the topic transfer; \textbf{3)} PD-DGD is better than PD-DGD(-$\mathcal{L}_{DA}$)/(-$\mathcal{L}_{Topic}$)/(-$\mathcal{L}_{DA}$,-$\mathcal{L}_{Topic}$). The reason is that our dialogue policy shares semantic overlaps with dialogue context so that the multitask learning schema in PD-DGD can leverage the policy prediction task to improve the dialogue understanding, and removing the policy prediction losses will cause declines; \textbf{4)} PD-DGD is better than PD-DGD(-Joint Learning), this result shows the Reinforce Learning process integrates the KS and RG and leverages the guiding signals from the generation to help KS; \textbf{5)} PD-DGD is better than PD-DGD(-Span Revision), which means the span revision method is useful for selecting the correct knowledge sentence. When removing this method, the performance on WoW/Holl-E/Doc2Dial of PD-DGD declines around 5/3/0.6 points, respectively. The performance on Doc2Dial suffers the least decline among three datasets. It is because the knowledge pieces in Doc2Dial are easier to locate (the KS accuracy on Doc2Dial is much higher than on WoW and Holl-E.) than the pieces in WoW and Holl-E. \textbf{6)} PD-DGD is better than PD-DGD(Sentence). Our span prediction and revision method is better than forcing the model to select one sentence. It is easier for the model to locate a text span containing the required knowledge rather than identify a whole sentence.

\begin{figure}[t]
\centering
\includegraphics[width=\linewidth]{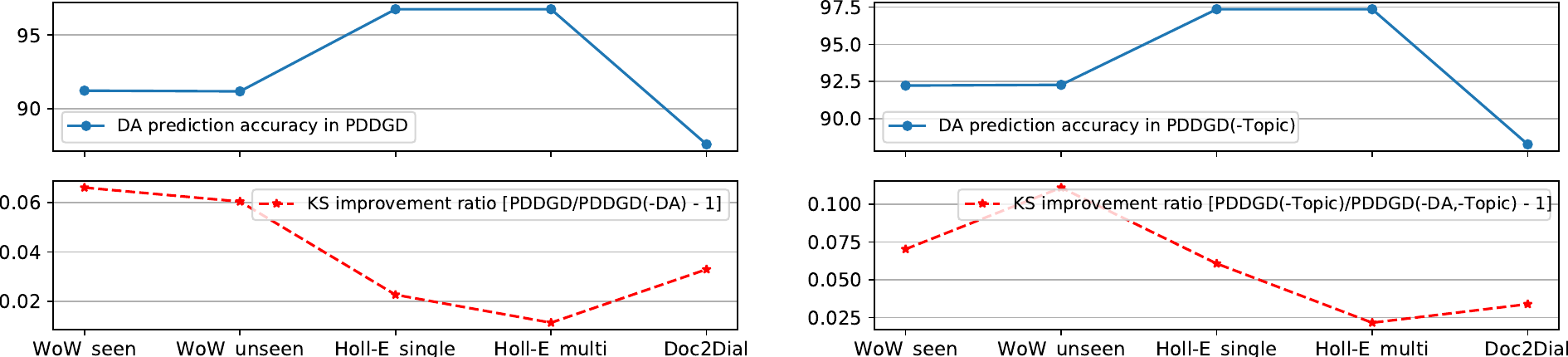}
\caption{Comparison of the DA prediction accuracy and the KS improvement ratio.}
\vspace{-0.5cm}
\label{DApredict}
\end{figure}

\subsubsection{DA Policy on Different Dialogue Data in KS}
\label{DAnaturalness}

In the experiments, we used two kinds of guiding signals on three different datasets. In this section, we study what is the effect when using the same kinds of guiding signals (DA) on different datasets. 

Figure \ref{DApredict} shows the effect of using DA. The top picture in the left part shows the DA prediction accuracy in the PD-DGD model. The bottom picture in the left part compares the KS results between PD-DGD(-DA) and PD-DGD by showing the KS improvement ratio, i.e. how much improvement in KS when adding DA information in PD-DGD(-DA). We use [PDDGD/PDDGD(-DA) - 1] to show this ratio in Figure \ref{DApredict}. 

We can see that the DA prediction accuracy is the highest on Holl-E, which means the DA mode in Holl-E is the simplest and easiest to predict. However, the KS improvement ratio is lowest (nearly 0\% improvement on Holl-E multiple Test set) on Holl-E, which means PD-DGD can hardly leverage DA to guide KS in Holl-E. The reason for these results lies in the dialogue mode. When constructing dialogues in Holl-E, the workers who act in the role of agents were asked to only add a few words before or after the selected knowledge sentence to form a response \cite{DBLP:conf/emnlp/MogheABK18}. Hence almost all agent turns are providing information and storing "inform" DA labels, which entails a less natural dialogue mode compared to WoW/Doc2Dial. 

Another piece of evidence supporting the conjecture, i.e. the dialogue mode in Holl-E is less natural than WoW/Doc2Dial, is the percentage of different DA labels. In Table \ref{trainDAdata}, the "inform" label ratio is only 65.35\% for the real human conversational data. Considering the three datasets we used are manually constructed by following given rules, it is reasonable that they are not as natural as the conversational data we used to train the DA tagger. In Table \ref{DA}, when using the DA tagger to annotate the three datasets, the "inform" label ratio is 79.01\%/79.09\%/84.01\% for WoW/Doc2Dial/Holl-E, respectively. A higher ratio of "inform" means a more unbalanced DA label and a more simple/unnatural dialogue mode. The Holl-E is the least unnatural one among the three. 

The top picture in the right part of Figure \ref{DApredict} shows the DA prediction accuracy in PD-DGD(-Topic) model where we only use DA as the dialogue policy\footnote{The DA prediction accuracy in PD-DGD(-Topic) is around 1\% percent higher than that in PD-DGD. It is reasonable since the PD-DGD(-Topic) only performs DA prediction and the PD-DGD performs both DA/topic-intent prediction.}. The bottom picture in the right part compares the KS results between PD-DGD(-DA,-Topic) and PD-DGD(-Topic) by showing the KS improvement ratio when adding DA information in PD-DGD(-DA,-Topic). We use [PDDGD(-Topic)/PDDGD(-DA,-Topic) - 1] to show this ratio in Figure \ref{DApredict}. We can see the trend in the right part is similar to the left part of Figure \ref{DApredict}. When removing the Topic intent, our model shows a consistent effect on the use of DA.

To sum up, the experimental results in Figure \ref{DApredict} shows that our model leverages the semantic information of the DA labels and works better on a more natural dialogue such as WoW/Doc2Dial than on a less natural dialogue such as Holl-E. When comparing WoW and Doc2Dial, the workers who constructed Doc2Dial needed to consider the pre-given dialogue acts when constructing the dialogue. This restriction makes the dialogues in Doc2Dial less natural than in WoW. In the bottom two pictures of Figure \ref{DApredict}, the higher improvement ratio of WoW than Doc2Dial again verifies that our model works better in a more natural dialogue mode.

\begin{figure}[t]
\centering
\includegraphics[width=\linewidth]{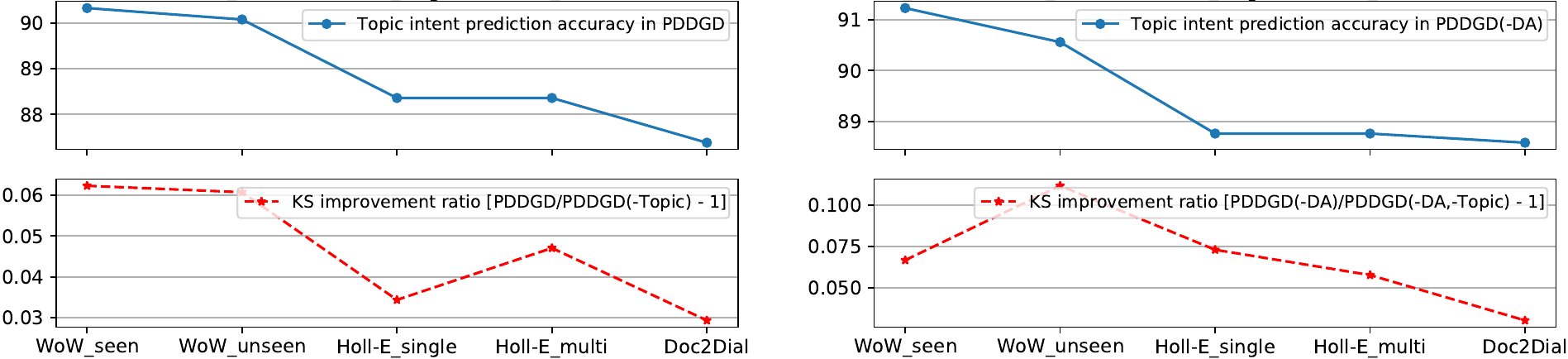}
\caption{Comparison of the Topic transfer intent prediction accuracy and the KS improvement ratio.}
\label{Topicpredict}
\end{figure}

\subsubsection{Topic Intent Policy on Different Dialogue Data in KS}
\label{documentTopic}

Similar to the analysis of DA, we provide the effect of using Topic intent in Figure \ref{Topicpredict}. The top picture on the left/right shows the Topic intent prediction accuracy in PD-DGD/PD-DGD(-DA), respectively. We can see that PD-DGD predicts topic intent better on WoW than on Holl-E and Doc2Dial. This is because the external knowledge in WoW is organized according to their shared topic entities, there is a clear gap between knowledge entries under different topics. In contrast, the external knowledge in Holl-E is organized according to its source (such as reviews or plots), and the external knowledge in Doc2Dial is organized according to paragraphs under one main topic (each paragraph focuses on one query and contains multiple knowledge entries that can answer the query). The topics in Holl-E and Doc2Dial are not as easy to distinguish since they have keyword overlaps. Hence, the topic transfer intents in the dialogue of Holl-E/Doc2Dial are not as easy to predict as in the dialogue of WoW.

The bottom picture on the left shows the KS improvement ratio between PD-DGD(-Topic) and PD-DGD. We use [PD-DGD/PD-DGD(-Topic) - 1] to show this ratio. The bottom picture on the right shows the KS improvement ratio between PD-DGD(-DA,-Topic) and PD-DGD(-DA). We use [PD-DGD(-DA)/PD-DGD(-DA,-Topic) - 1] to show this ratio. We can see that using topic intent information is more helpful on WoW and Holl-E than on Doc2Dial. The results show that the external knowledge in WoW (organized with shared entities) and Holl-E (organized with the same source) is easier for using topic transfer intent to select than in Doc2Dial (organized with paragraph).

Comparing the KS improvements between using DA and using Topic intent, we can see that the topic intent information is more helpful in the KS task. For example, the average KS improving ratio in Figure \ref{Topicpredict} is around 5.8\% and the ratio in Figure \ref{DApredict} is around 4.8\%. This is also reasonable since the Topic intent directly reflects the topic transfer and the knowledge utilization.

\subsubsection{Different or Larger Pre-trained Models}
\label{scaling}
\begin{table}[t]
\footnotesize
\begin{center}
\caption{Knowledge selection results (Hits@1) on the Doc2Dial Validation set. PD-DGD(BERT-base) is the base model to do the significant test. Values marked with * mean statistically significant with p$<$0.01.} 
\begin{tabular}{c |c | c | c}
\hline
Models & KS accuracy & Models & KS accuracy \\ 
\hline
PD-DGD(BERT-base)    & 59.6 & PD-DGD(RoBERTa-base)    &  61.4*\\
PD-DGD(BERT-large)    & 61.5*& PD-DGD(RoBERTa-large)    &  \textbf{62.1}*\\
\hline
\end{tabular}
\label{LMmore}
\end{center} 
\end{table}

In this paper, we use the BERT-base model as the policy planner for a fair comparison with the baselines. In this section, we test our dialogue policy on different or larger-size pre-trained models. We test with BERT-large (340m), RoBERTa-base (110m) \cite{DBLP:journals/corr/abs-1907-11692}, and RoBERTa-large (340m) beside BERT-base (110m). Table \ref{LMmore} shows the KS accuracy results on the Doc2Dial validation set. The results show that our dialogue policy can work with different pre-trained models and the performance increases when the model size became larger.

\begin{table}[t]
\caption{Generation results on the WoW Test seen/unseen sets. DukeNet is the base model to do the significant test for our models. Values marked with * mean statistically significant with p$<$0.01. "---" means the previous work did not report the result.}
\label{auto-wow}
\footnotesize
\begin{center} 
\begin{tabular}{l |c|c|c| c|c|c}
\hline
Models& PPL & F1(\%)  & BLEU-4(\%) & ROUGE-L & Dist-1(\%)  & Dist-2(\%) \\
\hline
TMN \cite{DBLP:conf/iclr/DinanRSFAW19}& 63.5 / 96.5 & 16.9 / 14.3    & 1.35 / 0.43 & 15.7 / 14.4  & 3.4 / 2.6   &  22.5 / 16.2\\ 
SKT \cite{DBLP:conf/iclr/KimAK20}              &  52.0 / 81.4  & 19.2 / 16.1   & 1.76 / 1.05  &  17.6 / 16.1 & 6.5 / 4.1 & 27.3 / 22.1\\
DukeNet \cite{DBLP:conf/sigir/MengRCSRTR20}& 48.3 / 69.4 & 19.3 / 17.1  & 2.43 / 1.68   &  18.5 / 17.1  & 6.6 / 5.0   & 27.4 / 22.0\\
CoLV \cite{DBLP:conf/emnlp/ZhanSCZ21}    & 39.6 / 54.3 &  --- / ---   & 2.85 / 2.12 	  & --- / ---  & --- / ---  & 29.7 / 20.1 \\
KnowledGPT  \cite{DBLP:conf/emnlp/ZhaoWXTZY20}  & 19.2 / 22.3  & 22.0 / 20.5  & 4.73 / 3.81   &  16.6 / 15.5 & 8.9 / 6.0  & 36.2 / 23.8  \\
\hline
GPT-2 (without policy)     & 23.5*/ 24.6* & 21.2*/ 20.3* & 4.43*/ 3.51*    & 16.2*/ 16.0* & 8.1*/ 5.8* & 35.3*/ 23.5* \\
GPT-2 (learned policy \cite{DBLP:conf/aaai/HazarikaNH22})    & 20.6*/ 20.9* & 23.1*/ 22.8* & 5.09*/ 4.27*    & 18.3*/ 17.4* & 9.0*/ 6.4* & 37.7*/ 24.1*\\
GPT-2 (stylistic policy \cite{DBLP:conf/acl-convai/SahaDS22})   & 22.9*/ 23.4* & 22.0*/ 21.2*   & 4.55*/ 3.62*    & 16.4*/ 16.1* & 8.2*/ 5.9* & 35.5*/ 23.7* \\
\hline
PD-DGD  & \textbf{19.1}*/ \textbf{19.5}*& \textbf{24.3}*/ \textbf{24.1}* & \textbf{5.52}*/ \textbf{4.91}*   & \textbf{18.9}*/ \textbf{17.8}* &	\textbf{10.5}*/ \textbf{7.0}*&  \textbf{38.6}*/ \textbf{25.8}* \\
\ (- DA)        & 20.2*/ 20.1* & 23.2*/ 23.0* & 5.13*/ 4.32*    & 18.7*/ 17.6* & 9.2*/ 6.6* & 38.0*/ 24.7* \\
\ (- Topic)         &20.7*/ 20.8* & 23.0*/ 22.5* & 5.05*/ 4.22*    & 18.0*/ 17.3* & 9.0*/ 6.5* & 37.8*/ 24.2*\\
\ (- DA, - Topic)  &22.6*/ 22.9* & 21.2*/ 20.5* & 4.63*/ 3.72*    & 16.7*/ 16.2* & 8.2*/ 5.8* & 35.5*/ 23.7* \\
\ (- Joint Learning) &20.5*/ 21.0* & 23.4*/ 23.1* & 5.08*/ 4.25*    & 18.1*/ 17.3* & 9.1*/ 6.5* & 37.6*/ 24.4* \\
\ (- Span Revision) &19.5*/ 19.5* & 23.9*/ 23.7* & 5.31*/ 4.66*    & 18.4*/ 17.4* & 9.5*/ 6.7* & 37.9*/ 24.9* \\
\ (Sentence)  &20.5*/ 20.7* & 23.3*/ 22.6* & 5.05*/ 4.21*    & 18.0*/ 17.2* & 9.1*/ 6.6* & 37.4*/ 24.3* \\ 
\ (ground truth)   & \textit{8.9*/ 9.4}* & \textit{38.2*/ 38.2}* &\textit{9.45*/ 8.16}*   & \textit{29.1*/  28.5}* & \textit{14.9*/ 11.1*} &\textit{42.9*/ 32.7}* \\
\hline
\end{tabular}
\end{center}
\end{table}

\subsubsection{The reasons that our dialogue policy helps the KS}
\label{whycontributes}
Besides the semantic overlap between the dialogue policy and the DGD task we introduced in Section \ref{introduction}, we further analyze why our dialogue policy (the utterance function and topic transfer intent) contributes to the knowledge selection process. One reason is from a semantic transfer perspective. The utterance functions and topic transfer intents are shared across different utterances, i.e. different utterances can have the same policy labels. The dialogue policy embeddings sever as "Messengers", transfer the shared semantic information among utterances with the same policy. The shared semantic information can help the DGD model to obtain a better representation of each utterance, consequently can help to learn a better context representation that is used to select knowledge. The other reason is that our policy contains semantic information that can narrow the range of knowledge candidates. For example, when the last utterance in the context has the "starting a new topic" intent label, the DGD model may learn to select the knowledge that does not exist in the context so as to improve the accuracy of knowledge selection.

\subsection{Response Generation (RG)}

\subsubsection{Comparing with models without dialogue policy}\label{no-policy-rg}
Table \ref{auto-wow} shows the generation results on WoW Test seen/unseen sets. We use PPL/F1/BLEU/ROUGE/Distinct results to compare our model and baselines. We can see that TMN could not compare with other baselines because of the lowest KS accuracy. SKT and DukeNet have fairly close performances. CoLV is better than SKT and DuckNet. These results are consistent with their performance on KS. On the other hand, KnowledGPT outperforms SKT/DukeNet/CoLV on all metrics. The reasons include \textbf{1)} KnowledGPT is more accurate in KS, especially on the Test Unseen set; \textbf{2)} the knowledge packed in the parameters of GPT-2 helps the generation. Finally, PD-DGD outperforms all baselines, including KnowledGPT. Table \ref{auto-holle} shows the generation results on Holl-E Test single/multi-reference and Doc2Dial validation sets. We can see that PD-DGD still outperforms all baselines.

\begin{table}[t]
\caption{Generation results on Holl-E Test single/multi-reference and Doc2Dial validation sets. DukeNet is the base model to do the significant test for our models. Values with * mean statistically significant with p$<$0.01. "---" means the previous work did not report the result, or the dataset does not have a manual policy.}
\label{auto-holle}
\footnotesize
\begin{center} 
\begin{tabular}{l |c|c|c| c|c|c}
\hline
Models& PPL & F1(\%)  & BLEU-4(\%) &  PPL & F1(\%)  & BLEU-4(\%)  \\
\hline
&\multicolumn{3}{c|}{{\centering Holl-E Test single/multi reference}} &\multicolumn{3}{c}{{\centering Doc2Dial validation}} \\
\hline
TMN    	& 66.5 / 90.1 & 15.9 / 14.1 & 6.77 / 8.98    &  72.5 &  23.3  & 7.11  \\ 
DukeNet  & 48.2 / 27.8 & 30.5 / 36.4 & 19.15 / 26.83   & 30.6 & 39.6 & 19.45  \\
CoLV       & 34.8 /  --- &   ---  /  ---  & 20.33  /  --- 	  & ---  & ---  & --- \\
\hline
\ GPT-2 (without policy) & 19.5*/ 18.0* & 37.1*/ 42.2* & 25.51*/ 32.13*    &  10.2*	 &   40.4*  & 22.15  \\
\ GPT-2 (manual policy \cite{DBLP:conf/emnlp/FengWGPJL20}) & ---  /  --- & ---  /  ---  & ---  /  --- 	   &  5.2*	 &   45.7*  & 24.11  \\
\ GPT-2 (learned policy \cite{DBLP:conf/aaai/HazarikaNH22}) & 18.3*/ 15.0* & 38.1*/ 43.4* & 25.76*/ 32.68*    &  9.3*	 &   41.9*  & 23.28  \\
\ GPT-2 (stylistic policy \cite{DBLP:conf/acl-convai/SahaDS22}) & 18.9*/ 16.0* & 37.3*/ 42.3* & 25.91*/ 33.20*    &  9.2*	 &   41.0*  & 23.25  \\
\hline
PD-DGD          & \textbf{15.8}*/ \textbf{11.2}         &  \textbf{39.4}*/ \textbf{45.2}*           & \textbf{28.80}*/ \textbf{35.61}*   &  \textbf{5.1}*	 &  \textbf{45.9}*  & \textbf{24.21*} \\
\ (- DA)          &16.6* / 11.5* & 38.9*  / 45.0* & 27.51*/ 34.38*    &  5.3*	 &  44.5*  & 23.65*  \\
\ (- Topic)       & 16.8*/ 12.3* & 38.8*/ 44.3* & 27.09*/ 33.87*   &  5.7* &  43.7*  & 23.24*  \\
\ (- DA, - Topic) & 18.7*/ 15.3* & 37.7*/ 42.8* & 25.55*/ 32.23*  &  8.3* &  41.4*  & 22.32*  \\
\ (- Joint Learning) & 16.7*/ 13.0 & 38.7*/ 44.6* & 27.22*/ 34.13* & 5.4* &   44.1*  & 23.55*  \\
\ (- Span Revision)        &16.5*/ 12.1* & 38.8*/ 45.0* & 27.46*/ 34.30*    &  5.4*	 &  44.6*  & 23.67*  \\
\ (Sentence)        &16.8*/ 12.2* & 38.6*/ 44.5* & 27.34*/ 33.91*    &  5.5*	 &  44.3*  & 23.59*  \\
\ (ground truth)         &\textit{2.3}*/\textit{2.3}* &\textit{76.6}*/\textit{76.6}* &\textit{72.80}*/\textit{72.80}*   &\textit{3.8}*	 &  \textit{51.1}*  & \textit{28.01}*    \\
\hline
\end{tabular}
\end{center}
\end{table}

\subsubsection{Comparing with other dialogue policy}\label{other-policy-rg}
Table \ref{auto-wow} and \ref{auto-holle} also show the generation results when using dialogue policies proposed by other researchers. On all datasets, GPT-2(learned policy) and GPT-2(stylistic policy) are better than GPT-2(without policy). These results show that \textbf{1)} the GPT-2 model can comprehend the semantic information in the dialogue policies; \textbf{2)} adding these semantic guiding signals can help to improve the generation quality. GPT-2(learned policy) has a fairly close performance with GPT-2(stylistic policy), this is also consistent with the KS results. The stylistic policy provides more kinds of semantic information than the learned policy but also introduces more noise than the learned policy. These noises make the semantic guiding signals in stylistic policy harder to learn. In contrast, GPT-2(manual policy), which uses a human-annotated, more accurate policy, is better than learned/stylistic policies on Doc2Dial. Finally, our PD-DGD is better than all baselines in three datasets, showing that our policy provides global instruction and helps our model to get better generation results.

\subsubsection{Ablation Study in RG}
\label{rg-ablation}
Table \ref{auto-wow} and \ref{auto-holle} also show the ablation study results in RG. On all datasets, \textbf{1)} PD-DGD is better than PD-DGD(-DA)/(-Topic)/(-DA,-Topic) and PD-DGD(-DA,-Topic) performs worst among all settings of PD-DGD. These results mean that DA and Topic intent are both useful in our method; \textbf{2)} We find that PD-DGD(-DA) is better than PD-DGD(-Topic), this is consistent with the KS results that removing topic intent will cause more declines than removing DA; \textbf{3)} The results of PD-DGD(-Joint Learning) are worse than PD-DGD, showing that removing the Reinforcement learning process also causes declines in response generation. The joint training helps to learn a better generator; \textbf{4)} The performances of PD-DGD(-Span Revision) and PD-DGD(Sentence) are worse than PD-DGD, which means better KS results entail better generation quality; \textbf{5)} PD-DGD(ground truth) is the upper bound of our method and it has much better results than PD-DGD, which means we still have a lot of work to do in the DGD task.

\begin{figure}[t]
\centering
\includegraphics[width=\linewidth]{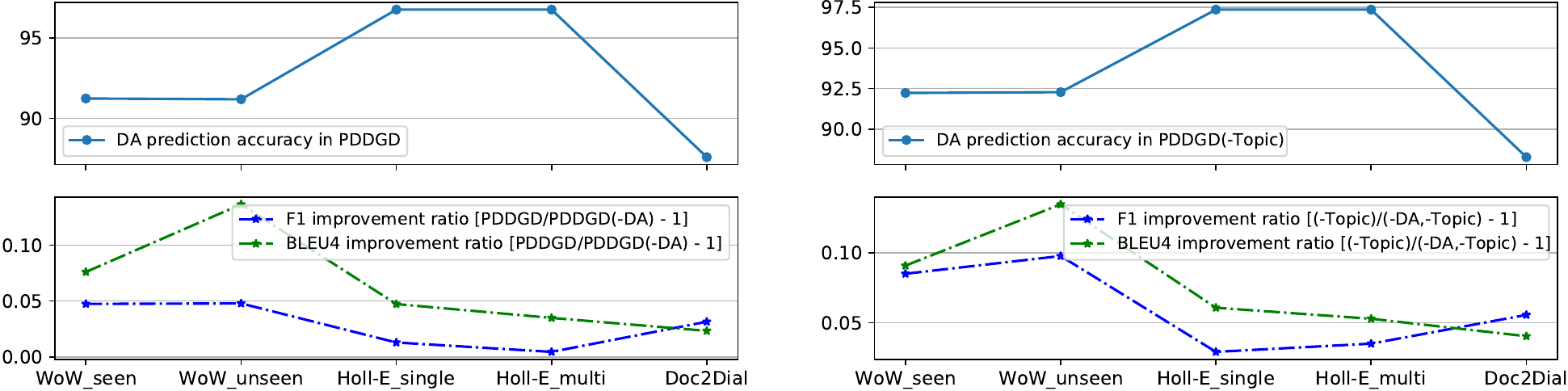}
\caption{Comparison of the DA prediction accuracy and the RG improvement ratio.}
\label{DARG}
\end{figure}

\subsubsection{DA policy on different dialogue data in RG}
\label{DAnaturalnessRG}
Similar to the analysis in KS, we study the effect of applying the DA policy on different dialogue data in RG. The top picture in the left part in Figure \ref{DARG} shows the DA prediction accuracy in the PD-DGD model. The bottom picture in the left part compares the F1 and BLEU4 results between PD-DGD(-DA) and PD-DGD by showing the improvement ratio. We use [PDDGD/PDDGD(-DA) - 1] to show this ratio. The top picture in the right part of Figure \ref{DApredict} shows the DA prediction accuracy in PD-DGD(-Topic) model where we only use DA as the dialogue policy. The bottom picture in the right part compares the F1/BLEU4 results between PD-DGD(-DA,-Topic) and PD-DGD(-Topic). We use [PDDGD(-Topic)/PDDGD(-DA,-Topic) - 1] to show the improvement ratio. We can see the trend of the improvement ratio in the left/right part is similar to the ratio in Figure \ref{DApredict}. These experimental results in Figure \ref{DARG} again show that our model works better on a more natural dialogue such as WoW than on a less natural dialogue such as Doc2Dial/Holl-E.

\begin{figure}[t]
\centering
\includegraphics[width=\linewidth]{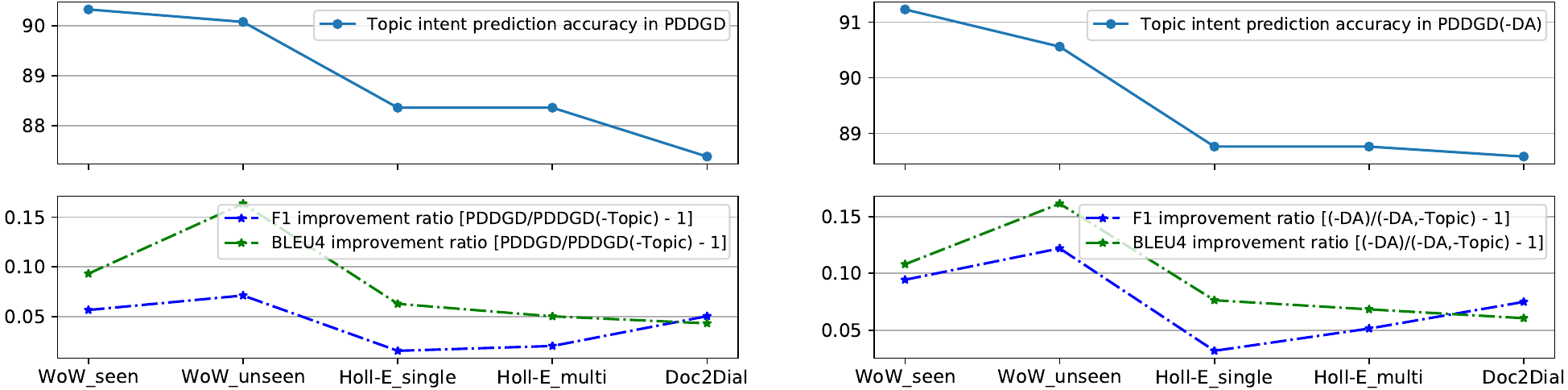}
\caption{Comparison of the Topic transfer intent prediction accuracy and the RG improvement ratio.}
\label{TopicRG}
\end{figure}

\subsubsection{Topic intent policy on different dialogue data in RG}
\label{TopicnaturalRG}
We also study the effect of using the Topic intent policy on different dialogue datasets in RG. The top picture on the left/right part in Figure \ref{TopicRG} shows the Topic intent prediction accuracy in PD-DGD/PD-DGD(-DA), respectively. The bottom picture on the left shows the KS improvement ratio between PD-DGD(-Topic) and PD-DGD. We use [PDDGD/PDDGD(-Topic) - 1] to show this ratio. The bottom picture on the right shows the KS improvement ratio between PD-DGD(-DA,-Topic) and PD-DGD(-DA). We use [PDDGD(-DA)/PDDGD(-DA,-Topic) - 1] to show this ratio. We can see that the F1 trend in Figure \ref{TopicRG} is similar to that in Figure \ref{Topicpredict}. In both Figures, the F1 improvement ratio is WoW>Holl-E>Doc2Dial. However, the BLEU4 trend in Figure \ref{TopicRG} is slightly different from that in Figure \ref{Topicpredict}. The BLEU4 ratio is WoW>Doc2Dial>Holl-E. This result indicates that we can improve the generation quality by improving the Generator's ability to use n-gram knowledge. However, the overall improvement on WoW is higher than that on Holl-E/Doc2Dial. This result again verifies that our method works better on a more natural dialogue.

Comparing the improvements between using DA and using Topic intent in RG, the average F1 improving ratio in Figure \ref{TopicRG} is around 8.8\% and the ratio in Figure \ref{DARG} is around 7.1\%, the average F1 improving ratio in Figure \ref{TopicRG} is around 5.9\% and the ratio in Figure \ref{DARG} is around 4.5\%. This is consistent with the analysis in KS that the topic intent information is more helpful than the DA information. It is reasonable since a higher improvement in KS will entail a higher improvement in RG.

\begin{table}[t]
\caption{Generation results on the WoW Test seen/unseen. PD-DGD (GPT-2 117m) is the base model to do the significant test for our models. Values with * mean statistically significant with p$<$0.01.}
\label{generationScaling}
\footnotesize
\begin{center} 
\begin{tabular}{l |c|c|c| c|c|c}
\hline
Models& PPL & F1(\%)  & BLEU-4(\%) & ROUGE-L & Dist-1(\%)  & Dist-2(\%) \\
\hline
PD-DGD(GPT-2 117m)  & 19.1 / 19.5& 24.3 / 24.1 & 5.52 / 4.91   & 18.9 / 17.8 &	10.5 / 7.0&  38.6 / 25.8 \\
PD-DGD(BART 139m) & 18.1*/ 18.5*& 25.2*/ 24.9* & 6.02*/ 5.41*   & 20.0*/ 19.1* &	10.6*/ 8.3*&  38.9*/ 26.7* \\
PD-DGD(GPT-2 345m) & 15.5*/ 15.4* & 26.4*/ 26.3* & 6.60*/ 6.51*    & 21.2*/ 20.8* & 12.1*/ 8.8* & 39.3*/ 28.5* \\
\hline
\end{tabular}
\end{center}
\end{table}

\subsubsection{Different or Larger Pre-trained Models}
\label{ScalingGen}

In this paper, we use the GPT-2 (117m) model as the Generator for a fair comparison with the baselines. In this section, we test our dialogue policy on different or larger-size pre-trained models. We test with GPT-2(345m) and BART(139m) \cite{DBLP:journals/corr/abs-1907-11692}. Table \ref{generationScaling} shows the generation results when replacing using GPT-2 or BART as Generator in PD-DGD. The results show that our dialogue policy can work with different pre-trained models and have a consistent improvement when the model size became larger.

\begin{table}[t]
\caption{Manual evaluation on the WoW Test unseen and Doc2Dial validation sets. "Flu."/"Coh."/"Rel." means "Fluency"/"Context Coherence"/"Knowledge Relevance", respectively. Values with * mean statistically significant with p$<$0.01. Values with \^\ mean statistically significant with p$<$0.05.}
\label{human}
\footnotesize
\begin{center} 
\begin{tabular}{l | cccc | l |cccc}
\hline
Models & Flu. &  Coh. & Rel. & Kappa & Models & Flu. &  Coh. & Rel. & Kappa \\
\hline
\multicolumn{5}{c|}{{\centering WoW Test unseen }} & \multicolumn{5}{c}{{\centering Doc2Dial validation }} \\
\hline
KnowledGPT & 1.67  & 1.50   & 1.61  & 0.68  &  DukeNet   & 1.63  & 1.42   & 1.53  & 0.62 \\
GPT-2(learned policy) & 1.65*    & 1.52\^      & 1.60* &  0.70 & GPT-2(manual policy) & 1.66*   & 1.56*      & 1.54\^ &  0.67 \\
\hline
PD-DGD  & 1.69*  & 1.62*   & 1.64* & 0.71 & PD-DGD  & 1.69*  & 1.61*   & 1.65* & 0.69\\
\hline
\end{tabular}
\end{center} 
\end{table}

\subsubsection{Manual Evaluation}
\label{manual-eval}
Since the automatic evaluation metrics alone may not be sufficient to reflect the dialogue quality, we show \textit{manual evaluation} results of RG in Table \ref{human}. The best performance models (DukeNet, KnowledGPT, GPT-2(learned policy), GPT-2(manual policy), and PD-DGD) are compared on Fluency / Context Coherence / Knowledge Relevance. We can see that PD-DGD is better than KnowledGPT/GPT-2(learned policy) on WoW and better than DukeNet/GPT-2(manual policy) on Doc2Dial, respectively. The overall inter-rater agreement measured by Fliess’ Kappa ranges from $0.62$ to $0.71$, indicating substantial agreement among the annotators. There are steady gaps between Kappa scores from the output of different models. We think the main reason for the steady gaps in the inter-annotator agreement is that our model can generate a response more related to the given knowledge. When evaluating, we provide the dialogue context, the response from different models, and ground truth knowledge to annotators. Since our model has a higher knowledge selection accuracy, the generated responses have a bigger chance to utilize the given ground truth knowledge, which yields a higher inter-annotator agreement. The manual evaluation further verifies that PD-DGD is a new state-of-the-art DGD model and that our dialogue policy can help the RG task.

\subsubsection{Case Study}
\label{case-study}

\begin{figure}[t]
\centering
\includegraphics[width=\linewidth]{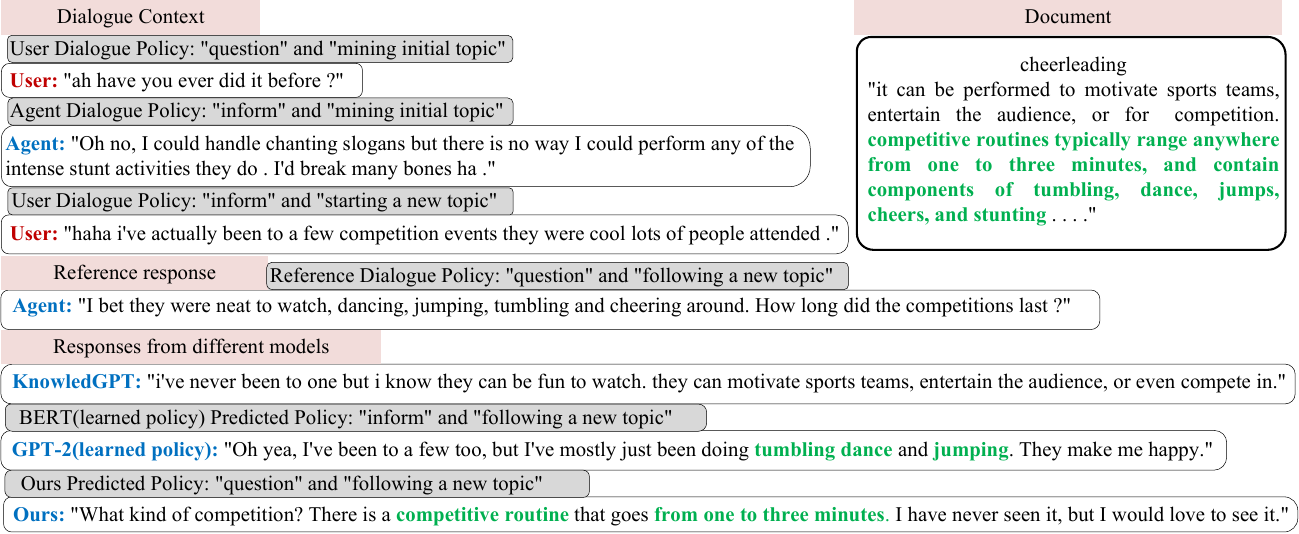}
\caption{Dialogue case from WoW Test-seen set. The case is randomly selected from the cases where our model selects the correct knowledge entry.}
\label{cheerleader}
\end{figure}

\begin{figure}[t]
\centering
\includegraphics[width=\linewidth]{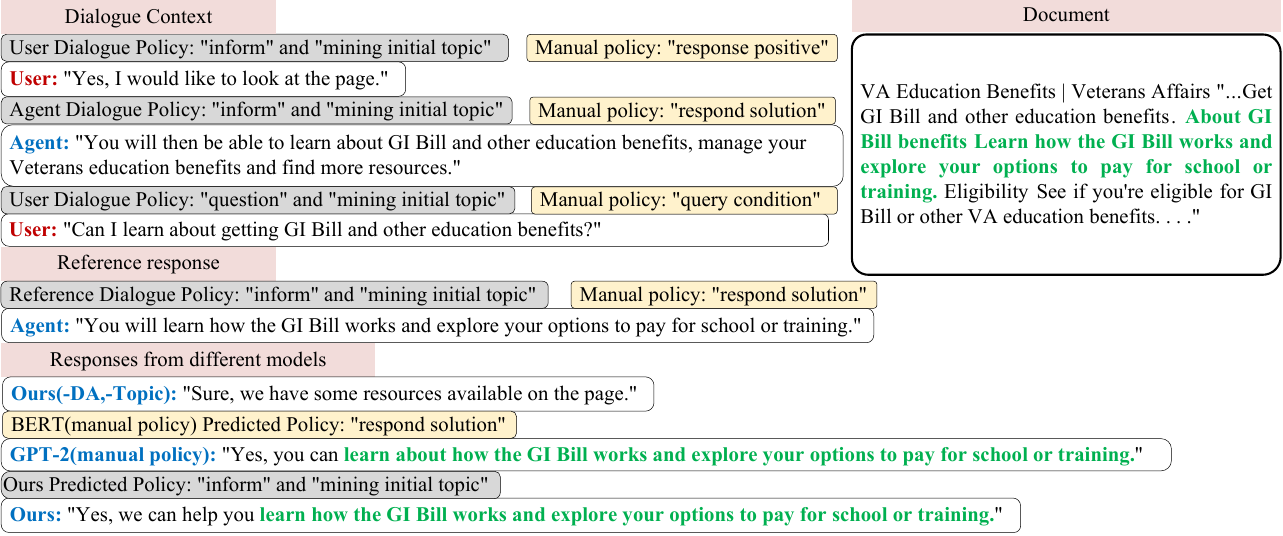}
\caption{Dialogue case from Doc2Dial Validation set. The case is randomly selected from the cases where our model selects the correct knowledge entry.}
\label{veteran}
\end{figure}

In Figure \ref{cheerleader}, we randomly select a dialogue case in WoW Test Seen set. The dialogue context (3 turns), corresponding policies, the golden response in the dataset, and part of the documents are presented. We show the responses of the three best performance models: KnowledGPT, GPT-2(learned policy), and PD-DGD. The knowledge entry and dialogue policy of GPT-2(learned policy) are predicted by BERT(learned policy). The ground-truth knowledge sentence in the document is bold and green. We can see that KnowledGPT fails to select the correct knowledge sentence, while BERT(learned policy) and PD-DGD succeed. The response from GPT-2(learned policy) is not context coherence, which shows that the RG is a challenging task even when the model selects the correct knowledge sentence. In contrast, with the help of the predicted dialogue policy "question" and "Following a new topic", the PD-DGD generates a coherent and informative response starting with a context-related question. 

In Figure \ref{veteran}, we randomly select a dialogue case in the Doc2Dial Validation set to compare different dialogue policies. GPT-2(manual policy) leverages the knowledge entry and dialogue policy predicted by BERT(manual policy) to generate the response. We can see BERT(manual policy) and our PD-DGD both predict the correct knowledge entry and dialogue policy. They generate similar responses to the reference one. PD-DGD(-DA,-Topic) is our model without any dialogue policy, it selects the wrong knowledge and generates a less informative reply. This case verifies that our dialogue policy is useful and shows that our policy is comparable with the expensive manual-annotated policy. 

\begin{figure}[t]
\centering
\includegraphics[width=\linewidth]{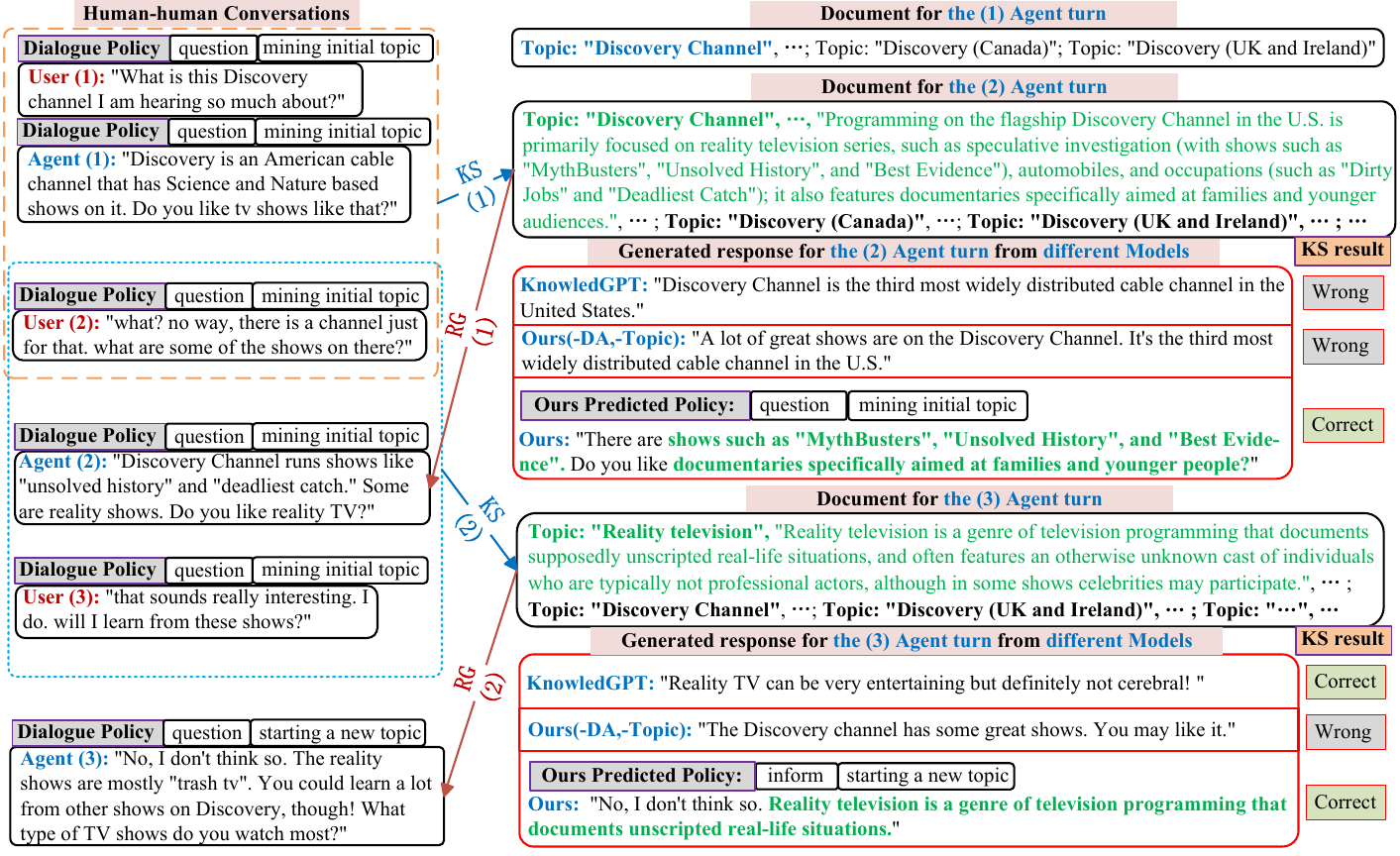}
\caption{Two KS/RG cases in a multi-turn dialogue from WoW Test-seen set. The KS(1) line means selecting knowledge with the first three turns (User(1), Agent(1), and User(2)). The RG(1) line means generating a response for Agent(2) turn with the selected knowledge. Similarly, the KS(2) means selecting knowledge with the User(2), Agent(2), and User(3) turns, RG(2) means generating response for Agent(3) turn. The case is randomly selected from the cases where our model selects the correct knowledge entry.}
\label{Discovery}
\end{figure}

In Figure \ref{Discovery}, we present another dialogue example (including two cases) in WoW Test seen set that PD-DGD selects the correct knowledge entries. The dialogue context, the annotated/predicted dialogue policies, the golden responses, and the ground-truth knowledge entries (bold and green) are presented. We show the responses of the KnowledGPT, PD-DGD(-DA,-Topic), and PD-DGD. For agent turn (2), KnowledGPT and PD-DGD(-DA,-Topic) fail to select the correct knowledge while PD-DGD succeeds. The response from PD-DGD is context coherence and reflects the "question" DA. In Agent's turn (3), PD-DGD predicts a wrong DA and fails to produce a topic-related question as the reference agent's turn (3). However, PD-DGD predicts the correct topic transfer intent "starting a new topic" and selects the correct knowledge. Benefiting from the correct KS results, both PD-DGD and KnowledGPT generate an informative and coherent response. In contrast, PD-DGD(-DA,-Topic) still fails on KS and gives an irrelevant response. These two cases again show that the dialogue policy helps the KS and RG in the DGD task. There are also cases that our model performs not as well as the examples we selected, such as selecting wrong knowledge entries or generating incoherent responses. We will further improve the KS and RG performance of PD-DGD in the future.

\subsubsection{Experiments for automatically learning the bias weight}
\label{automaticweight}
We tried several methods to treat the bias weight vector as trainable parameters. The bias weight represents the effect of the policy on the response. To learn the bias weight during the training, we need to first have valid policy embeddings for the generation model, so that the model can leverage the semantic information of the policy embeddings to learn the bias weight. We tested with two different methods to obtain the policy embeddings. The first directly encodes the text description of the policy as \textbf{text embeddings}. The second is \textbf{learned embeddings} which is the average of a set of sentence embeddings. These sentences have the same policy label and their embeddings are obtained by the encoder of the generation module.

Notice that the bias weight is equal to a gate that assigns different weights to the knowledge and the context. We test with two gates to obtain the bias weight through the generator in Figure 2. The first gate is as follows:

\begin{align}
\alpha &= \mu([\textbf{e}_p; \textbf{e}_t]W_{bias} ),
\end{align}

where $\mu$ is sigmoid function, [;] is the concatenation operation, $\textbf{e}_p$ is the policy embedding, $\textbf{e}_t$ is the output of the self-attention layer when the $t$-th token of the response is generated, $W_{bias}$ is parameters to be learned. The $\alpha$ is the weight for the knowledge and (1-$\alpha$) is the weight for dialogue context. For example, if the $\alpha$ = 0.8, the bias weight vector $\textbf{b}$ is equal to [$4_1$,$4_2$,...,$4_m$,$1_1$,$1_2$,...,$1_n$]. The first method is denoted by (policy only). The second gate is:

\begin{align}
\alpha &= \mu([\textbf{e}_p; \textbf{e}_{K_i, \textbf{C}}; \textbf{e}_t]W_{bias} ),
\end{align}

where $\textbf{e}_{K_i, \textbf{C}}$ is the encoding results of the knowledge and the dialogue context. The symbols are the same as in Section \ref{thegenerator}. The second method is denoted by (policy+).

\begin{table}
\footnotesize
\caption{Generation experiments on the WoW Test seen set when automatically learning the bias weight.}
\label{upperboundtest}
\centering
\begin{tabular}{ l  | c c c c c}
\hline
Setting &\textbf{PPL} &\textbf{F1} & \textbf{BLEU4} &\textbf{ROUGE-L}  &\textbf{Dist-2} \\
\hline
Text embeddings (policy only) & 12.8 & 36.6 & 9.10 & 26.6 & 39.2 \\
Text embeddings (policy+)       & 16.3 & 33.6 & 8.66 & 25.5 & 36.2\\
\hline
Learned embeddings (policy only) &12.6 & 36.8 & 9.21 &27.6 &39.5 \\
Learned embeddings (policy+)        &15.1& 35.1 & 8.82 &26.1 &36.7\\
\hline
Empirical Weight &\textbf{8.9} &\textbf{38.2}  &\textbf{9.45} &\textbf{29.1}  &\textbf{42.9} \\
\hline
\end{tabular}
\end{table}

Table \ref{upperboundtest} shows the generation results of the above settings on the WoW Test seen set. The input to the generator is model-annotated dialogue policy, ground-truth knowledge sentences, and dialogue context. The generation results can be seen as the upper bound of different settings. We compare these results with the upper bound of the empirical weight method and they are not good enough. These results indicate that introducing the policy embedding makes the generator more difficult to train. We may need a different generator structure to directly leverage the semantic information of our policy for a generation. In contrast, the policy planner can easily leverage our policy since the task space in the policy planner is much smaller than the generator. The policy planner only predicts start/end positions in a document or predicts the next policies in a few categories while the task space of the generator is the entire vocabulary.

Compared to the automatic learning method, the empirical method can directly reflect the effect of different policies and is easier to train. Meanwhile, the generator has a higher upper bound according to our tests. This advantage is the reason we use this method. However, there may be other methods or model structures that can automatically learn the bias weights. We consider exploring these methods or models as future work.

\begin{table}
\footnotesize
\caption{Generation experiments on the WoW Test seen set to decide the bias weight. "other DA" means "question/directive/commissive". "starting"/"mining"/"following" is short for "starting a new topic"/"mining the initial topic"/"following a new topic", respectively.}
\label{decidebias}
\centering
\begin{tabular}{ c  c | c  c  c | c c c c c}
\hline
"inform" &"other DA" &"starting" &"mining" &"following" &\textbf{PPL} &\textbf{F1} & \textbf{BLEU4} &\textbf{ROUGE-L}  &\textbf{Dist-2} \\
\hline
1 &  1 & 0 & 0 & 0 & \textbf{8.8} & 38.0 & 9.39 &28.3  &42.3 \\
2 &  2 & 0 & 0 & 0 & 8.9 & 38.1 & 9.41 &28.4  &42.4 \\
3 &  3 & 0 & 0 & 0 & 9.5 & 38.1 & 9.43 &28.5  &42.6 \\
4 &  4 & 0 & 0 & 0 & 9.9 & 38.0 & 9.42 &28.5  &42.7 \\
5 &  5 & 0 & 0 & 0 & 12.0 & 37.8 & 9.40 &28.4  &42.7 \\
\hline
1 &  0 & 2 & 1 & 0 & 8.9 & 37.9 & 9.38 &28.1  &42.3 \\
1 &  1 & 2 & 1 & 0 & 8.9 & 38.0 & 9.40 &28.4  &42.5 \\
2 &  1 & 2 & 1 & 0 & 8.9 &\textbf{38.2}  &\textbf{9.45} &\textbf{29.1}  &\textbf{42.9} \\
2 &  2 & 2 & 1 & 0 & 9.3 & 37.6 & 9.41 & 28.6 & 42.5 \\
3 &  2 & 2 & 1 & 0 & 9.8 & 37.6 & 9.40 & 28.5 & 42.5 \\
\hline
1 & 0 & 3 & 2 & 1  & 9.1 & 38.1 & 9.42 &28.9  &42.7 \\
1 & 1 & 3 & 2 & 1  & 9.4 & 38.0 & 9.41 &28.6  &42.5 \\
2 & 1 & 3 & 2 & 1  & 9.7 & 37.9 & 9.42 &28.5  &42.4 \\
2 & 2 & 3 & 2 & 1 & 10.7 & 38.0 & 9.39 & 28.4 &42.4 \\
3 & 2 & 3 & 2 & 1 & 11.9 & 37.8 & 9.39 & 28.3 &42.4 \\
\hline
2 & 1 & 1 & 1 & 0 & 9.1 & 38.0 & 9.41 &28.4  &42.4 \\
2 & 1 & 1 & 1 & 1 & 9.3 & 38.0 & 9.41 &28.5  &42.5 \\
2 & 1 & 2 & 1 & 1 & 9.3 & 37.8 & 9.42 &28.6  &42.6 \\
2 & 1 & 2 & 2 & 1 & 9.7 & 37.7 & 9.40 &28.5  &42.4 \\
2 & 1 & 2 & 2 & 2 & 9.9 & 37.7 & 9.39 &28.5  &42.3 \\
\hline
\end{tabular}
\end{table}

\subsubsection{Experiments for deciding the bias weight}
\label{decide-bias-weight}
In Section \ref{thegenerator}, we analyze how our dialogue policy may affect knowledge utilization and we expect the bias weight to reflect this effect. The four DA labels can be categorized into two levels ("inform" and "question/directive/commissive") and the topic transfer intent has three levels. In Table \ref{bias-examples}, the dialogue context weight is fixed with 1 and we determine the knowledge weights by experiments. The experimental results for the WoW Test seen set are shown in Table \ref{decidebias}. We select the bias weight with the best performance generator when given the ground truth knowledge, i.e. the (ground truth) setting in Table \ref{auto-wow}. There are four groups in Table \ref{decidebias}. The first group tests the range of the weight for knowledge. We can see that the performance drops when the weight is larger than 4. The second and the third groups fix the weights of Topic transfer intent labels and test different DA weights. Similar to the first group, the PPL decreases when the weight increases. The results of other metrics are close except the third row of the second group has the best scores. The fourth group further fixes the weight of the best performance DA weights and tests different Topic transfer intent weights. We can see that the best performance is still the third row of the second group. Experiments on the other datasets show a similar trend to Table \ref{decidebias}. Finally, we choose a weight of 2 for "inform" and a weight of 1 for "question/directive/commissive". The weight for "starting a new topic"/"mining the initial topic"/"following a new topic" is 2/1/0, respectively. It is possible that we can learn a better weight by dividing the policies into more levels or testing with more fine-grained weights. However, the current weight is good enough to verify the effectiveness of the bias weight method and we leave the more fine-grained exploring as future work.

\section{Conclusions}
In this paper, we propose a policy-driven model for knowledge selection (KS) and response generation (RG) in Document-grounded Dialogue (DGD). Different from previous work, we propose a dialogue policy to assist the dialogue modeling in both KS and RG, providing a global interpretation for the DGD task. Our dialogue policy contains utterance functions and topic transfer intents, which are closely related to the dialogue understanding and knowledge utilization of the DGD task. For KS and RG, our Policy-driven DGD (PD-DGD) model leverages different mechanisms to efficiently integrate the semantic information of the dialogue policy. Experiments on three public datasets show that our method not only outperforms DGD models without dialogue policy but also surpasses other dialogue policies proposed by previous researchers. We give a detailed analysis of why leveraging our policy could help these tasks and show the difference when using the same policy on different dialogue data. Future works include but are not limited to 1) we would like to test the effect of our policy-driven model on more DGD datasets, such as those without ground-truth knowledge labels or those in other morphology languages \cite{DBLP:conf/sigir/RenLSTCRR21}; 2) we would like to test whether our dialogue policy can work together with external document clues such as reconstructing document into a graphic \cite{DBLP:conf/coling/XuZFKN22}.

\section*{Ethics Statement and Limitations}
The datasets we used in this paper are all English data from previously published papers and are all publicly available. We did not make any changes to the data so there are no data ethical problems in this research. The effectiveness of the method proposed in this paper has only been verified on English datasets. Whether this method works in other morphology languages requires further verification.

\begin{acks}
This research is supported by the Science and Technology Innovation 2030 Major Project of China (No. 2021ZD0113302), National Natural Science Foundation of China (No. 62076081, No. 61772153 and No. 61936010) and Nature Scientific Foundation of Heilongjiang Province(YQ2021F006).
\end{acks}

\bibliographystyle{ACM-Reference-Format}
\bibliography{sample-base}

\end{document}